\documentclass[sigconf]{acmart}
\usepackage{booktabs} 
\usepackage{tabularx}
\usepackage{graphicx}
\usepackage{url}
\usepackage{mdwlist}
\usepackage{subcaption}
\usepackage{hyperref}
\hypersetup{colorlinks=false,linkcolor=blue,urlcolor=blue,citecolor=red}
\usepackage{amsfonts}
\usepackage[labelfont=bf,textfont=bf]{caption}
\usepackage{multirow}
\usepackage{array}
\usepackage[linesnumbered,ruled]{algorithm2e}
\SetKwComment{Comment}{$\triangleright$\ }{}
\usepackage{xcolor}
\setcopyright{rightsretained}
\usepackage{amssymb}
\usepackage[flushleft]{threeparttable}
\usepackage{enumitem}
\usepackage{balance}

\newcommand{\ie}{\emph{i.e., }}
\newcommand{\eg}{\emph{e.g., }}
\newcommand{\etal}{\emph{et al.}}

\newcommand{\wrt}{\emph{w.r.t. }}
\newcommand{\aka}{\emph{a.k.a. }}
\newcommand{\cf}{\emph{cf. }}

\begin{document}

\title{Neural Factorization Machines for Sparse Predictive Analytics}
\titlenote{NExT research is supported by the National Research Foundation, Prime Minister's Office, Singapore under its IRC@SG Funding Initiative.}

\author{Xiangnan He}
\affiliation{%
	\institution{School of Computing}
	\streetaddress{National University of Singapore}
	\city{Singapore} 
	\postcode{117417}
}
\email{dcshex@nus.edu.sg}

\author{Tat-Seng Chua}
\affiliation{%
	\institution{School of Computing}
	\streetaddress{National University of Singapore}
	\city{Singapore} 
	\postcode{117417}
}
\email{dcscts@nus.edu.sg}


\copyrightyear{2017} 
\acmYear{2017} 
\setcopyright{acmcopyright}
\acmConference{SIGIR'17}{}{August 7--11, 2017, Shinjuku, Tokyo, Japan}
\acmPrice{15.00}\acmDOI{http://dx.doi.org/10.1145/3077136.3080777}
\acmISBN{978-1-4503-5022-8/17/08}
\fancyhead{}

\begin{abstract}
Many predictive tasks of web applications need to model categorical variables, such as user IDs and demographics like genders and occupations. To apply standard machine learning techniques, these categorical predictors are always converted to a set of binary features via one-hot encoding, making the resultant feature vector highly sparse. To learn from such sparse data effectively, it is crucial to account for the interactions between features. 

\textit{Factorization Machines} (FMs) are a popular solution for efficiently using the second-order feature interactions. However, FM models feature interactions in a linear way, which can be insufficient for capturing the non-linear and complex inherent structure of real-world data. While deep neural networks have recently been applied to learn non-linear feature interactions in industry, such as the \textit{Wide\&Deep} by Google and \textit{DeepCross} by Microsoft, the deep structure meanwhile makes them difficult to train. 

In this paper, we propose a novel model \textit{Neural Factorization Machine}~(NFM) for prediction under sparse settings. NFM seamlessly combines the linearity of FM in modelling second-order feature interactions and the non-linearity of neural network in modelling higher-order feature interactions.
Conceptually, NFM is more expressive than FM since FM can be seen as a special case of NFM without hidden layers. Empirical results on two regression tasks show that with one hidden layer only, NFM significantly outperforms FM with a $7.3\%$ relative improvement. Compared to the recent deep learning methods Wide\&Deep and DeepCross, 
our NFM uses a shallower structure but offers better performance, 
being much easier to train and tune in practice. 




\end{abstract}

\maketitle

\section{Introduction}
\label{sec:introduction}

Predictive analytics is one of the most important techniques for many information retrieval (IR) and data mining (DM) tasks, ranging from recommendation systems~\cite{iCD,He:WWW2017}, targeted advertising~\cite{fieldFM}, to search ranking~\cite{xiong2017learning,hong2015learning}, visual analysis~\cite{wang2015visual}, and event detection~\cite{zhang2016geoburst}. 
Typically, a predictive task is formulated as estimating a function that maps predictor variables to some target, for example, real valued target for regression and categorical target for classification.
Distinct from continuous predictor variables that are naturally found in images and audios, such as raw features, the predictor variables for web applications are mostly discrete and categorical.
For example, in online advertising, we need to predict \textit{how likely}~(target) a \textit{user}~(first predictor variable) of a particular \textit{occupation}~(second predictor variable) will click on an \textit{ad}~(third predictor variable). To build predictive models with these categorical predictor variables, a common solution is to convert them to a set of binary features~(\aka feature vector) via one-hot encoding~\cite{iCD,WideDeep,He:WWW2017,DeepCross,fastFM}. Thereafter, standard machine learning (ML) techniques such as logistic regression and support vector machines can be applied. 

Depending on the possible values of categorical predictor variables, the generated feature vector can be of very high dimension but sparse. To build effective ML models with such sparse data, it is crucial to account for the interactions between features~\cite{blondel2016polynomial,DeepCross,EM}. Many successful solutions in both industry and academia largely rely on manually crafting combinatorial features~\cite{WideDeep}, \ie constructing new features by combining multiple predictor variables, also known as \textit{cross features}. 
For example, we can cross variable $occupation=\{banker,  doctor\}$ with $gender=\{M, F\}$ and get a new $occupation\_gender = \{banker\_M, banker\_F, doctor\_M, doctor\_F\}$.
It is well known that top data scientists are usually masters of crafting combinatorial features, which play a key role in their winning formulas~\cite{DeepCross}.
However, the power of such features comes at a high cost,
since it requires heavy engineering efforts and useful domain knowledge to design effective features. Thus these solutions can be difficult to generalize to new problems or domains. 

Instead of augmenting feature vectors manually, another solution is to design a ML model to learn feature interactions from raw data automatically. A popular approach is factorization machines (FMs)~\cite{FM}, which embeds features into a latent space and models the interactions between features via inner product of their embedding vectors. While FM has yielded great promise\footnote{https://securityintelligence.com/factorization-machines-a-new-way-of-looking-at\-machine-learning} in many prediction tasks~\cite{iCD,chen2016context,fieldFM,ImportanceFM}, we argue that its performance can be limited by its linearity, as well as the modelling of pairwise (\ie second-order) feature interactions only. 
Specifically, for real-world data that have complex and non-linear underlying structure, FM may not be expressive enough.
Although higher-order FMs have been proposed~\cite{FM}, they still belong to the family of linear models and are claimed to be difficult to estimate~\cite{libFM}. Moreover, they are known to have only marginal improvements over FM, which we suspect the reason might be due to the modelling of higher-order interactions in a linear way.  



In this work, we propose a novel model for sparse data prediction named \textit{Neural Factorization Machines} (NFMs), which enhances FMs by modelling higher-order and non-linear feature interactions. 
By devising a new operation in neural network modelling --- \textit{Bilinear Interaction} (Bi-Interaction) pooling --- we subsume FM under the neural network framework for the first time. 
Through stacking non-linear layers above the Bi-Interaction layer, we are able to deepen the shallow linear FM, modelling higher-order and non-linear feature interactions effectively to improve FM's expressiveness. 
In contrast to traditional deep learning methods that simply concatenate~\cite{WideDeep,DeepCross,zhang2016deep} or average~\cite{He:WWW2017,Wang:2015} embedding vectors in the low level, our use of Bi-Interaction pooling encodes more informative feature interactions, greatly facilitating the following ``deep'' layers to learn meaningful information. 
We conduct extensive experiments on two public benchmarks for context-aware prediction and personalized tag recommendation.
With one hidden layer only, our NFM significantly outperforms FM~(the official LibFM~\cite{libFM} implementation) with a $7.3\%$ improvement. 
Compared to the state-of-the-art deep learning methods --- the 3-layer Wide\&Deep~\cite{WideDeep} and 10-layer DeepCross~\cite{DeepCross} --- our 1-layer NFM shows consistent improvements with a much simpler structure and fewer model parameters. Our implementation is available at: \url{https://github.com/hexiangnan/neural_factorization_machine}.

The main contributions of this work are summarized as follows. 
\begin{enumerate}[leftmargin=*]
\item To the best of our knowledge, we are the first to introduce the Bi-Interaction pooling operation in neural network modelling, and present a new neural network view for FM. 
\item Based on this new view, we develop a novel NFM model to deepen FM under the neural network framework for learning higher-order and non-linear feature interactions. 
\item We conduct extensive experiments on two real-world tasks to study the Bi-Interaction pooling and NFM model, demonstrating the effectiveness of NFM and great promise in using neural networks for prediction under sparse settings. 
\end{enumerate}

\section{Modelling Feature Interactions}
\label{sec:related}

Due to the large space of combinatorial features, traditional solutions typically rely on manual feature engineering efforts or feature selection techniques like boosted decision trees to select important feature interactions. 
One limitation of such solutions is that they cannot generalize to combinatorial features that have not appeared in the training data. 
In recent years, embedding-based methods become increasingly popular, which try to learn feature interactions from raw data~\cite{fastFM,ImportanceFM,CoFM,zhang2016deep,WideDeep,DeepCross}. 
By embedding high-dimensional sparse features into a low-dimensional latent space, the model can generalize to unseen feature combinations. 
Regardless of domain, we can categorize the approaches into two types: 1) factorization machine-based linear models, and 2) neural network-based non-linear models. In what follows, we shortly recapitulate the two representative techniques. 

\subsection{Factorization Machines}
Factorization machines are originally proposed for collaborative recommendation~\cite{FM,fastFM}. 
Given a real valued feature vector $\textbf{x}\in\mathbb{R}^n$, FM estimates the target by modelling all interactions between each pair of features via factorized interaction parameters:
\begin{equation}
	\hat{y}_{FM}(\textbf{x}) = w_0 + \sum_{i=1}^n w_i x_i + \sum_{i=1}^n\sum_{j=i+1}^n \textbf{v}_i^T \textbf{v}_j\cdot x_i x_j,
\end{equation}
where $w_0$ is the global bias, $w_i$ models the interaction of the $i$-th feature to the target.
The $\textbf{v}_i^T \textbf{v}_j$ term denotes the factorized interaction, where $\textbf{v}_i\in \mathbb{R}^k$ denotes the embedding vector for feature $i$, and $k$ is the size of embedding vector, also termed as \textit{number of latent factors} in literature. Note that due to the coefficient $x_i x_j$, only interactions of non-zero features are considered. 


One main power of FM stems from its generality --- in contrast to matrix factorization (MF) that models the relation of two entities only~\cite{fastMF}, FM is a general predictor working with any real valued feature vector for supervised learning. Clearly, it enhances linear/logistic regression (LR) using the second-order factorized interactions between features. 
By specifying input features, Rendle~\cite{FM} showed that FM can mimic many specific factorization models such as the standard MF, parallel factor analysis, and SVD++~\cite{SVD++},
Owing to such genericity, FM has been recognized as one of the most effective embedding methods for sparse data prediction. It has been successfully applied to many predictive tasks, ranging from 
online advertising~\cite{fieldFM}, 
microblog retrieval~\cite{Qiang:2013}, to open relation extraction~\cite{petroni2015core}.  

\subsubsection{\textbf{Expressiveness Limitation of FM}}
Despite effectiveness, we point out that FM still belongs to the family of (multivariate) linear models. In other words, the predicted target $\hat{y}(\textbf{x})$ is linear \wrt each model parameter~\cite{libFM}. Formally, for each model parameter $\theta\in \{w_0, \{w_i\}, \{v_{if}\}\}$, we can have $\hat{y}(\textbf{x}) = g + h \theta$, where $g$ and $h$ are expressions independent of $\theta$. 
Unfortunately, the underlying structure of real-world data is often highly non-linear and cannot be accurately approximated by linear models~\cite{genzel2016mathematical}. 
As such, FM may suffer from insufficient representation ability for modelling real data with complex inherent structure and regularities. 

In terms of methodology, many variants~\cite{CoFM,ImportanceFM,fieldFM,AFM} of FM have been developed. For example, Hong \etal~\cite{CoFM} proposed Co-FMs to learn from multi-view data; Oentaryo~\etal~\cite{ImportanceFM} encoded prior knowledge of features to FM by designing a hierarchical regularizer; and Lin \etal~\cite{fieldFM}~proposed field-aware FM, which learned multiple embedding vectors for a feature to differentiate its interaction with features of different fields.
More recently, Xiao \etal~\cite{AFM} proposed attentional FM, using an attention network~\cite{Attentive_CF,xiong2017learning} to learn the importance of each feature interaction.
However, these variants are all linear extensions of FM and model the second-order feature interactions only. As such, they can suffer from the same expressiveness issue for modelling real-world data. 

In this work, we contribute improvements on the expressiveness of FM by endowing it the ability of non-linearity modelling. The idea is to perform non-linear transformation on the latent space of the second-order feature interactions; meanwhile, higher-order feature interactions can be captured. 

\subsection{Deep Neural Networks}
\label{sec:dnn}
\begin{figure}[t]
	\centering
	\begin{subfigure}[b]{0.24\textwidth}
		\centering
		\includegraphics[width=\textwidth]{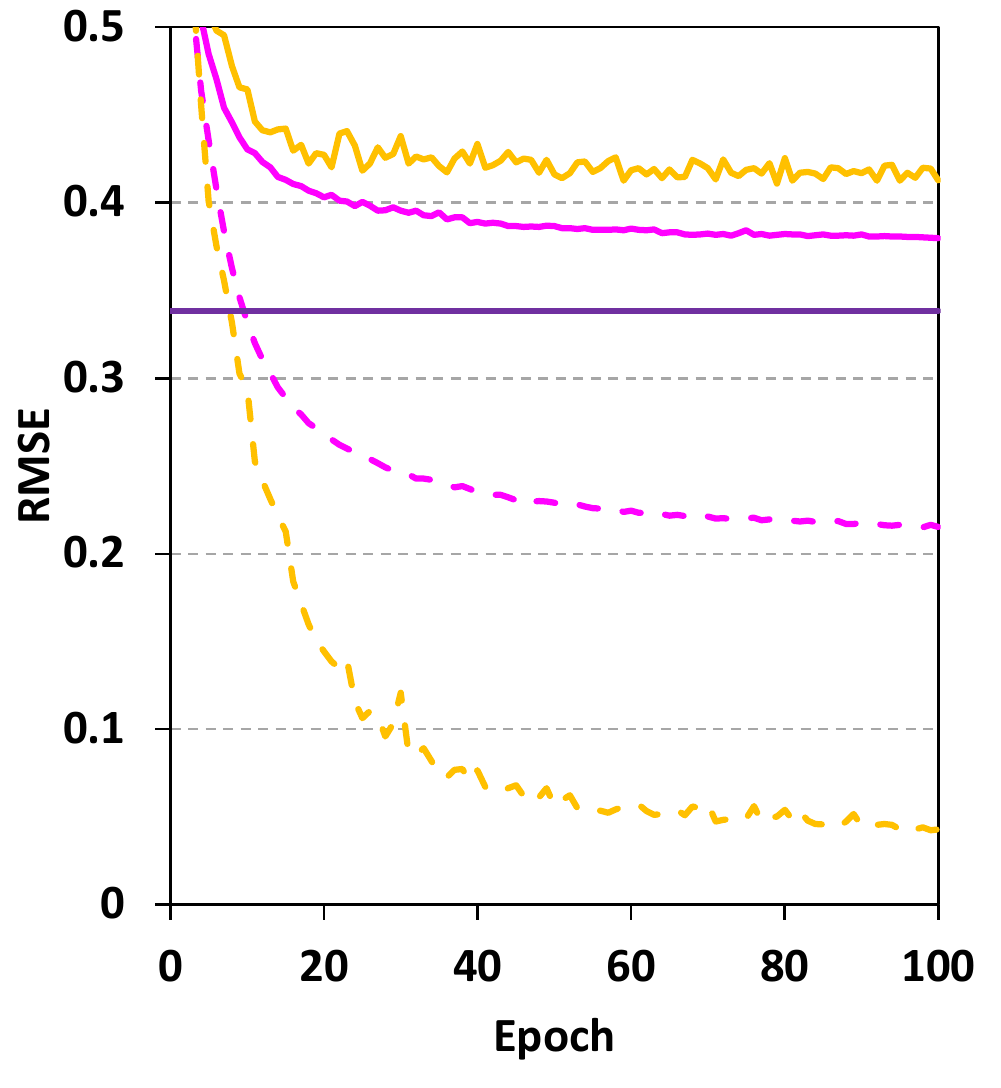}
		\vspace{-15pt}
		\caption{Random initialization}
		\label{fig:deep_difficulty}
	\end{subfigure} \hspace{-7pt}
	\begin{subfigure}[b]{0.24\textwidth}
		\centering
		\includegraphics[width=\textwidth]{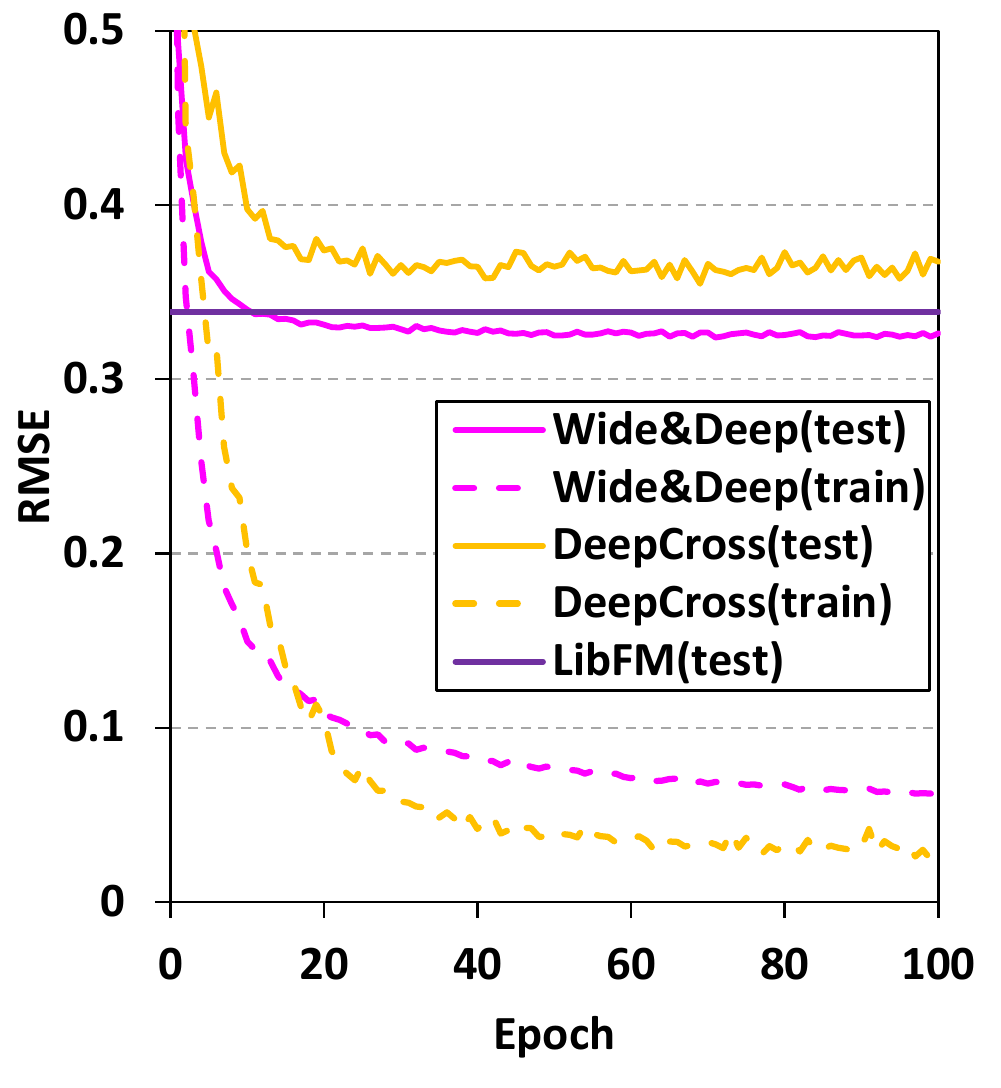}
		\vspace{-15pt}
		\caption{FM as pre-training}
		\label{fig:deep_difficulty_pretrain}
	\end{subfigure} \hspace{-7pt}
	\vspace{-5pt}
	\caption{Training and test error of each epoch of Wide\&Deep and DeepCross on Frappe. Random initialization of embeddings leads to poor performance, worse than LibFM (a). Initializing using the embeddings learned by FM improves the performance significantly (b). This reveals  optimization difficulties for training deep neural networks.}
	\vspace{-15pt}
	\label{fig:difficulty}
\end{figure}

In recent five years, deep neural networks (DNNs) have achieved immense success and have been widely used on speech recognition, computer vision and natural language processing. However, the use of DNNs is not as widespread among the IR and DM community. In our view, we think one reason might be that most data of IR and DM tasks are naturally sparse, such as user behaviors, documents/queries and various features converted from categorical variables. Although DNNs have exhibited strong ability to learn patterns from dense data~\cite{cvpr16best}, the use of DNNs on sparse data has received less scrutiny, and it is unclear how to employ DNNs for effectively learning feature interactions under sparse settings. 

Until very recently, some work~\cite{He:WWW2017,WideDeep,DeepCross,zhang2016deep,DeepCTR} started to explore DNNs for some scenarios of sparse predictive analytics. 
In \cite{He:WWW2017}, He \etal~presented a neural collaborative filtering (NCF) framework to learn interactions between users and items.
Later, the NCF framework was extended to model attribute interactions for attribute-aware CF~\cite{silkroad}.
However, their methods are only applicable to learn interactions between two entities and do not directly support the general setting of supervised learning. 
Zhang \etal~\cite{zhang2016deep} developed a FM-supported Neural Network (FNN), which uses the feature embeddings learned by FM to initialize DNNs.  
Cheng \etal~\cite{WideDeep} proposed Wide\&Deep for App recommendation, where the deep part is a multi-layer perceptron (MLP) on the concatenation of feature embedding vectors to learn feature interactions.
Shan~\etal~\cite{DeepCross} proposed DeepCross for ads prediction, which shares a similar framework with Wide\&Deep by replacing the MLP with the state-of-the-art residual network~\cite{cvpr16best}.
 
\subsubsection{\textbf{Optimization Difficulties of DNN}}
\label{ss:difficulty_dnn}
It is worthwhile to mention the common structure of these neural network-based approaches, \ie stacking multiple layers above the concatenation of embedding vectors to learn feature interactions. 
The expectation is that the multiple layers can learn combinatorial features of arbitrary orders in an implicit way~\cite{DeepCross}.
However, we find a key weakness of such an architecture is that simply concatenating feature embedding vectors carries too little information about feature interactions in the low level. 
An empirical evidence is from He \etal's recent work \cite{He:WWW2017}, which shows that simply concatenating user and item embedding vectors leads to very poor results for collaborative filtering.
To remedy this, one has to rely on the following deep layers to learn meaningful interaction function. 
While it is claimed that multiple non-linear layers are able to learn feature interactions well~\cite{WideDeep,DeepCross}, such a deep architecture can be difficult to optimize in practice due to the notorious problems of vanishing/exploding gradients, overfitting, degradation, among others~\cite{cvpr16best}.

To demonstrate optimization difficulties of DNNs empirically, we plot the training and test error of each epoch of Wide\&Deep and DeepCross on the Frappe data in Figure~\ref{fig:difficulty}. We use the same architecture and parameters as reported in their papers, where Wide\&Deep applies a 3-layer tower structure and DeepCross uses a 10-layer residual network with successively decreasing hidden units. 
From Figure~\ref{fig:deep_difficulty}, we can see that training the two models from scratch leads to a performance much worse than the shallow FM model. 
For Wide\&Deep, the training error is relatively high, which is likely because of the degradation problem~\cite{cvpr16best}.
For DeepCross, the very low training error but high test error implies that the model is overfitting.
Inspired by FNN~\cite{zhang2016deep}, we further explore the use of feature embeddings learned by FM to initialize DNNs, which can be seen as a pre-training step.
As can be seen from Figure~\ref{fig:deep_difficulty_pretrain}, both models achieve much better performance~(over $11\%$ improvements). For Wide\&Deep, the degradation problem is well addressed, evidenced by the much lower training error, and the test performance is better than LibFM. However for the 10-layer DeepCross, it still suffers from severe overfitting and underperforms LibFM. 
These relatively negative results reveal optimization difficulties for training DNNs. 

In this work, we present a new paradigm of neural networks for sparse data prediction. Instead of concatenating feature embedding vectors, we propose a new Bi-Interaction operation that models the second-order feature interactions. This results in a much more informative representation in the low level, greatly helping the subsequent non-linear layers to learn higher-order interactions. 




\section{Neural Factorization Machines}
\label{sec:method}
We first present the NFM model that unifies the strengths of FMs and neural networks for sparse data modelling.
We then discuss the learning procedure and how to employ some useful techniques in neural networks --- dropout and batch normalization ---  for NFM. 

\subsection{The NFM Model}
\begin{figure}[t]
	\centering
	\includegraphics[width=0.45\textwidth]{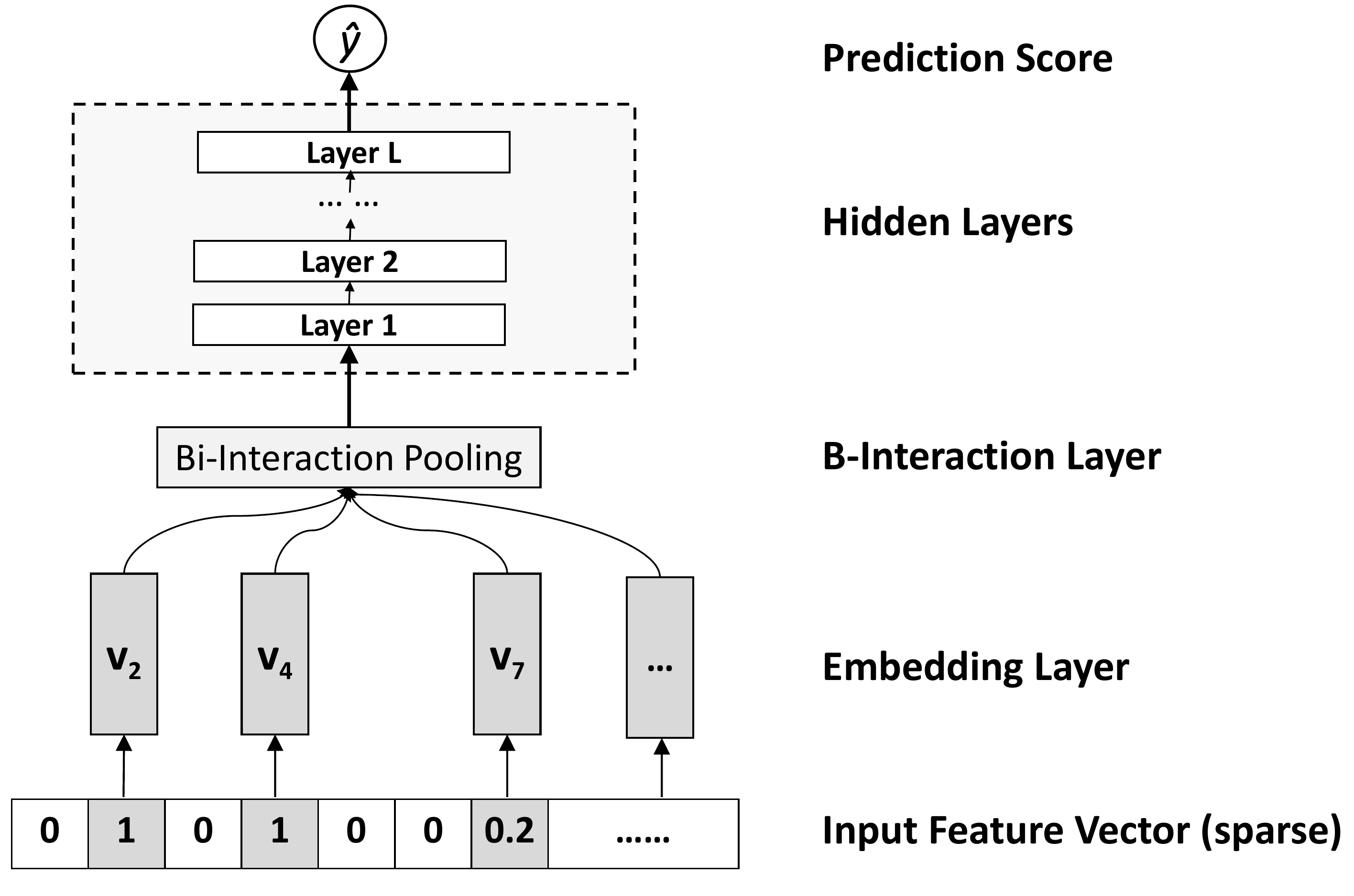}
	\vspace{-5pt}
	\caption{Neural Factorization Machines model (the first-order linear regression part is not shown for clarity).}
	\vspace{-10pt}
	\label{fig:nfm_model}
\end{figure}

Similar to factorization machine, NFM is a general machine learner working with any real valued feature vector.
Given a sparse vector $\textbf{x}\in \mathbb{R}^n$ as input, where a feature value $x_i = 0$ means the $i$-th feature does not exist in the instance, NFM estimates the target as:
\begin{equation}
	\hat{y}_{NFM}(\textbf{x}) = w_0 + \sum_{i=1}^n w_i x_i + f(\textbf{x}),
\end{equation}
where the first and second terms are the linear regression part similar to that for FM, which models global bias of data and weight of features. The third term $f(\textbf{x})$ is the core component of NFM for modelling feature interactions, which is a multi-layered feed-forward neural network as shown in Figure~\ref{fig:nfm_model}. In what follows, we elaborate the design of $f(\textbf{x})$ layer by layer. \\\vspace{-5pt}

\textbf{Embedding Layer.} The embedding layer is a fully connected layer that projects each feature to a dense vector representation. Formally, let $\textbf{v}_i\in \mathbb{R}^k$ be the embedding vector for the $i$-th feature. After embedding, we obtain a set of embedding vectors $\mathcal{V}_x = \{x_1 \textbf{v}_1, ..., x_n \textbf{v}_n\}$ to represent the input feature vector $\textbf{x}$. 
Owing to sparse representation of $\textbf{x}$, we only need to include the embedding vectors for non-zero features, \ie $\mathcal{V}_x = \{x_i \textbf{v}_i \}$ where $x_i\neq 0$.
Note that we have rescaled an embedding vector by its input feature value, rather than simply an embedding table lookup, so as to account for the real valued features~\cite{FM}. 

\textbf{Bi-Interaction Layer.} We then feed the embedding set $\mathcal{V}_x$ into the Bi-Interaction layer, which is a pooling operation that converts a set of embedding vectors to one vector:
\begin{equation}
\label{eq:BI}
f_{BI}(\mathcal{V}_{x}) = \sum_{i=1}^n \sum_{j=i+1}^n x_i\textbf{v}_i\odot x_j\textbf{v}_j,
\end{equation}
where $\odot$ denotes the element-wise product of two vectors, that is, $(\textbf{v}_i \odot \textbf{v}_j)_k = v_{ik}v_{jk}$. 
Clearly, the output of Bi-Interaction pooling is a $k$-dimension vector that encodes the second-order interactions between features in the embedding space. 

It is worth pointing out that our proposal of Bi-Interaction pooling does not introduce extra model parameter, and more importantly, it can be efficiently computed in linear time. 
This property is the same with average/max pooling and concatenation that are rather simple but commonly used in neural network  approaches~\cite{Wang:2015,He:WWW2017,DeepCross}.
To show the linear time complexity of evaluating Bi-Interaction pooling, we reformulate Equation~(\ref{eq:BI}) as:
\begin{equation}
\label{eq:BI2}
f_{BI}(\mathcal{V}_{x}) = \frac{1}{2} \left[( \sum_{i=1}^n x_i\textbf{v}_i)^2 - \sum_{i=1}^n (x_i\textbf{v}_i)^2 \right],
\end{equation}
where we use the symbol $\textbf{v}^2$ to denote $\textbf{v}\odot\textbf{v}$. By considering the sparsity of $\textbf{x}$, we can actually perform Bi-Interaction pooling in $O(k N_x)$ time, where $N_x$ denotes the number of non-zero entries in $\textbf{x}$. This is a very appealing property, meaning that the benefit of Bi-Interaction pooling in modelling pairwise feature interactions does not involve any additional cost. 

\textbf{Hidden Layers}. 
Above the Bi-Interaction pooling layer is a stack of fully connected layers, which are capable of learning higher-order interactions between features~\cite{DeepCross}. Formally, the definition of fully connected layers is as follows:
\begin{equation}
\begin{aligned}
	\textbf{z}_1 &= \sigma_1 (\textbf{W}_1 f_{BI}(\mathcal{V}_x) + \textbf{b}_1 ), \\
	\textbf{z}_2 &= \sigma_2 (\textbf{W}_2 \textbf{z}_1 + \textbf{b}_2), \\
	&\quad\quad ... ... \\
	\textbf{z}_L &= \sigma_L (\textbf{W}_L \textbf{z}_{L-1} + \textbf{b}_L),
\end{aligned}
\end{equation}
where $L$ denotes the number of hidden layers, $\textbf{W}_l$, $\textbf{b}_l$ and $\sigma_l$ denote the weight matrix, bias vector and activation function for the $l$-th layer, respectively. By specifying non-linear activation functions, such as sigmoid, hyperbolic tangent (tanh), and Rectifier (ReLU), we allow the model to learn higher-order feature interactions in a non-linear way. 
This is advantageous to existing methods for higher-order interaction learning, such as higher-Order FM~\cite{blondel2016higher} and Exponential Machines~\cite{EM}, which only support the learning of higher-order interactions in a linear way. 
As for the structure of fully connected layers (\ie size of each layer), one can freely choose tower~\cite{WideDeep,He:WWW2017}, constant~\cite{DeepCross}, and diamond~\cite{zhang2016deep}, among others. 

\textbf{Prediction Layer.} At last, the output vector of the last hidden layer $\textbf{z}_L$ is transformed to the final prediction score:
\begin{equation}
	f(\textbf{x}) = \textbf{h}^T \textbf{z}_L, 
\end{equation}
where vector $\textbf{h}$ denotes the neuron weights of the prediction layer. \\\vspace{-5pt}

\noindent To summarize, we give the formulation NFM's predictive model as:
\begin{equation}
\begin{aligned}
	\hat{y}_{NFM}(\textbf{x}) &= w_0 + \sum_{i=1}^n w_i x_i  \\
	&+ \textbf{h}^T \sigma_L(\textbf{W}_L(...\sigma_1(\textbf{W}_1 f_{BI}(\mathcal{V}_x) + \textbf{b}_1)  ... ) + \textbf{b}_L),
\end{aligned}
\end{equation}
with all model parameters $\Theta = \{w_0, \{w_i, \textbf{v}_i\}, \textbf{h}, \{\textbf{W}_l, \textbf{b}_l\} \}$. Compared to FM, the additional model parameters of NFM are mainly $\{\textbf{W}_l, \textbf{b}_l\}$, which are used for learning higher-order interactions between features. In remainder of this subsection, we first show how NFM generalizes FM and discuss the connection of NFM between existing deep learning methods; we then analyze the time complexity of evaluating NFM model. 

\subsubsection{\textbf{NFM Generalizes FM}} 
\label{ss:nfm_generalizes_fm}
FM is a shallow and linear model, which can be seen as a special case of NFM with no hidden layer. 
To show this, we set $L$ to zero and directly project the output of Bi-Interaction pooling to prediction score. 
We term this simplified model as NFM-0, which is formulated as:
\begin{equation}
\begin{aligned}
	\hat{y}_{NFM-0} &= w_0 + \sum_{i=1}^n w_i x_i + \textbf{h}^T \sum_{i=1}^n \sum_{j=i+1}^n x_i\textbf{v}_i\odot x_j\textbf{v}_j \\
	& = w_0 + \sum_{i=1}^n w_i x_i + \sum_{i=1}^n \sum_{j=i+1}^n \sum_{f=1}^k h_f v_{if} v_{jf}\cdot x_i x_j.
\end{aligned}
\end{equation}
As can be seen, by fixing $\textbf{h}$ to a constant vector of $(1,...,1)$, we can exactly recover the FM model\footnote{Note that for NFM-0, a trainable $\textbf{h}$ can not improve the expressiveness of FM, since its impact on prediction can be absorbed into feature embeddings.}. 

It is worth pointing out that, to our knowledge, this is the first time FM has been expressed under the neural network framework.
While a recent work by Blondel \etal~\cite{blondel2016polynomial} has unified FM and Polynomial network via kernelization, their kernel view of FM only provides a new training algorithm and does not provide insight for improving the FM model.
Our new view of FM is highly instructive and provides more insight for improving FM. 
Particularly, we allow the use of various neural network techniques on FM to improve its learning and generalization ability.
For example, we can use dropout~\cite{srivastava2014dropout} --- a well-known technique in deep learning community to prevent overfitting --- on the Bi-Interaction layer as a way to regularize FM, which we find can be even more effective than the conventional $L_2$ regularization (see Figure~\ref{fig:dropout_reg} of Section~\ref{sec:experiments}).

\subsubsection{\textbf{Relation to Wide\&Deep and DeepCross}}
NFM has a similar multi-layered neural architecture with several existing deep learning solutions~\cite{WideDeep,DeepCross,He:WWW2017}. 
The key difference is in the Bi-Interaction pooling component, which is uniquely used in NFM. 
Specifically, if we replace the Bi-Interaction pooling with concatenation and apply a tower-structure MLP (or residual units~\cite{cvpr16best}) to hidden layers, we can recover the Wide\&Deep (or DeepCross) model.
An obvious limitation of the concatenation operation is that it does not account for any interaction between features. 
As such, these deep learning approaches have to rely entirely on the following deep layers to learn meaningful feature interactions, which unfortunately can be difficult to train in practice (\cf Section~\ref{ss:difficulty_dnn}). 
Our use of Bi-Interaction pooling captures second-order feature interactions in the low level, which is more informative than the concatenation operation. 
This greatly facilitates the subsequent hidden layers of NFM to learn useful higher-order feature interactions in a much easier way. 

\subsubsection{\textbf{Time Complexity Analysis}}
\label{ss:time_complexity}
We have shown in Equation~(\ref{eq:BI2}) that Bi-Interaction pooling can be efficiently computed in $O(k N_x)$ time, which is the same as FM. Then the additional costs are caused by the hidden layers. 
For hidden layer $l$, the matrix-vector multiplication is the main operation which can be done in $O(d_{l-1} d_l)$, where $d_l$ denotes the dimension of the $l$-th hidden layer and $d_0=k$. The prediction layer only involves inner product of two vectors, for which the complexity is $O(d_L)$. 
As such, the overall time complexity for evaluating a NFM model is $O(k N_x + \sum_{l=1}^L d_{l-1}d_l )$, which is the same as that of Wide\&Deep and DeepCross. 

\subsection{Learning}
NFM can be applied to a variety of prediction tasks, including regression, classification and ranking. 
To estimate model parameters of NFM, we need to specify an objective function to optimize. 
For regression, a commonly used objective function is the squared loss:
\begin{equation}
\label{eq:squared_loss}
	L_{reg} = \sum_{\textbf{x}\in \mathcal{X}} (\hat{y}(\textbf{x}) - y(\textbf{x}))^2,
\end{equation}
where $\mathcal{X}$ denotes the set of instances for training, and $y(\textbf{x})$ denotes the target of instance $\textbf{x}$. The regularization terms are optional and omitted here, since we found that some techniques in neural network modelling such as dropout can well prevent NFM from overfitting. 
For classification task, we can optimize the hinge loss or log loss~\cite{He:WWW2017}. For ranking task, we can optimize pairwise personalized ranking loss~\cite{BPR,silkroad} or contrastive max-margin loss~\cite{zhang2013attribute}. 
In this work, we focus on the regression task and optimize the squared loss of Equation~(\ref{eq:squared_loss}). The optimization for ranking/classification tasks can be done in the same way. 

Stochastic gradient descent (SGD) is a universal solver for optimizing neural network models. 
It iteratively updates the parameters until convergence. In each time, it randomly selects a training instance $\textbf{x}$, updating each model parameter towards the direction of its negative gradient:
\begin{equation}
	\theta = \theta - \eta\cdot 2 (\hat{y}(\textbf{x}) - y(\textbf{x})) \frac{d \hat{y}(\textbf{x})}{d \theta},
\end{equation}
where $\theta\in \Theta$ is a trainable model parameter, and $\eta>0$ is the learning rate that controls the step size of gradient descent. 
As NFM is a multi-layered neural network model, the gradient of $\hat{y}(\textbf{x})$ \wrt to each model parameter can be derived using the chain rule. 
Here we give only the differentiation of the Bi-Interaction pooling layer, since other layers are just standard operations in neural network modelling and have been widely implemented in ML toolkits like TensorFlow and Keras.
\begin{equation}
	\frac{d f_{BI}(\mathcal{V}_{x})}{d \textbf{v}_i} = (\sum_{j=1}^n x_j \textbf{v}_j) x_i - x_i^2 \textbf{v}_i =  \sum_{j=1, j\neq i}^n x_i x_j \textbf{v}_j.
\end{equation} 
As such, for end-to-end neural methods, upon plugging in the Bi-Interaction pooling layer, they can still be learned end-to-end.
To leverage the vectorization and parallelism speedup of modern computing platforms, mini-batch SGD is more widely used in practice, which samples a batch of training instances and updates model parameters based on the batch.
In our implementation, we use mini-batch Adagrad~\cite{Adagrad} as the optimizer, rather than the vanilla SGD. Its main advantage is that the learning rate can be self adapted during the training phase, which eases the pain of choosing a proper learning rate and leads to faster convergence than the vanilla SGD. 

\subsubsection{\textbf{Dropout}} 
While neural network models have strong representation ability, they are also easy to overfit the training data. Dropout~\cite{srivastava2014dropout} is a regularization technique for neural networks to prevent overfitting. The idea is to randomly drop neurons (along with their connections) of the neural network during training. That is, in each parameter update, only part of the model parameters that contributes to the prediction of $\hat{y}(\textbf{x})$ will be updated. 
Through this process, it can prevent complex co-adaptations of neurons on training data. 
It is important to note that in the testing phase, dropout must be disabled and the whole network is used for estimating $\hat{y}(\textbf{x})$. As such, dropout can also be seen as performing model averaging with smaller neural networks~\cite{srivastava2014dropout}. 

In NFM, to avoid feature embeddings co-adapt to each other and overfit the data, we propose to adopt dropout on the Bi-Interaction layer. Specifically, after obtaining $f_{BI}(\mathcal{V}_{x})$ which is a $k$-dimensional vector of latent factors, we randomly drop $\rho$ percent of latent factors, where $\rho$ is termed as the dropout ratio. 
Since NFM with no hidden layer degrades to the FM model, it can be seen as a new way to regularize FM. 
Moreover, we also apply dropout on each hidden layer of NFM to prevent the learning of higher-order feature interactions from co-adaptations and overfitting. 

\subsubsection{\textbf{Batch Normalization}}
One difficulty of training multi-layered neural networks is caused by the fact of covariance shift~\cite{ioffe2015batch}. It means that the distribution of each layer's inputs changes during training, as the parameters of the previous layers change.
As a result, the later layer needs to adapt to these changes (which are often noisy) when updating its parameters, which adversely slows down the training. To address the problem, Ioffe and Szegedy~\cite{ioffe2015batch} proposed \textit{batch normalization} (BN), which normalizes layer inputs to a zero-mean unit-variance Gaussian distribution for each training mini-batch. It has been shown that BN leads to faster convergence and better performance in several computer vision tasks~\cite{cvpr16best,zhang2016play}.

Formally, let the input vector to a layer be $\textbf{x}_i\in \mathbb{R}^d$ and all input vectors to the layer of the mini-batch be $\mathcal{B}=\{\textbf{x}_i\}$, 
then BN normalizes $\textbf{x}_i$ as:
\begin{equation}
	BN(\textbf{x}_i) = \gamma\odot(\frac{\textbf{x}_i - \mu_B}{\sigma_B}) + \beta,
\end{equation}
where $\mu_B = \frac{1}{|\mathcal{B}|}\sum_{i\in \mathcal{B}} \textbf{x}_i$ denotes the mini-batch mean, $\sigma_B^2 = \frac{1}{|\mathcal{B}|}\sum_{i\in \mathcal{B}} (\textbf{x}_i - \mu_B)^2 $ denotes the mini-batch variance, and $\gamma$ and $\beta$ are trainable parameters (vectors) to scale and shift the normalized value to restore the representation power of the network. 
Note that in testing, BN also needs to be applied, where $\mu_B$ and $\sigma_B$ are estimated from the whole training data. 

In NFM, to avoid the update of feature embeddings changing the input distribution to hidden layers or prediction layer, we perform BN on the output of the Bi-Interaction pooling. For each successive hidden layer, the BN is also applied. 


\section{Experiments}
\label{sec:experiments}
As the key contributions of this work are on Bi-Interaction pooling in neural network modelling and the design of NFM for sparse data prediction, we conduct experiments to answer the following research questions:
\begin{itemize}
	\item[\textbf{RQ1}] Can Bi-Interaction pooling effectively capture the second-order feature interactions? How does dropout and batch normalization work for Bi-Interaction pooling? 
	\item[\textbf{RQ2}] Are hidden layers of NFM useful for capturing higher-order interactions between features and improving the expressiveness of FM? 
	\item[\textbf{RQ3}] How does NFM perform as compared to higher-order FM and the state-of-the-art deep learning methods Wide\&Deep and DeepCross?
\end{itemize}

In what follows, we first present the experimental settings, followed
by answering the above research questions one by one.

\subsection{Experimental Settings}
\subsubsection{\textbf{Datasets}} We experimented with two publicly accessible datasets: Frappe\footnote{\url{http://baltrunas.info/research-menu/frappe}} and MovieLens\footnote{\url{http://grouplens.org/datasets/movielens/latest}}: \\\vspace{-5pt}

\textbf{1. Frappe}.  Frapp\'e is a context-aware app discovery tool. This dataset is constructed by Baltrunas \etal~\cite{FrappeData}. It contains $96,203$ app usage logs of users under different contexts. Besides user ID and app ID, each log contains 8 context variables, including weather, city and daytime (\eg morning or afternoon). We converted each log (\ie user ID, app ID and all context variables) to a feature vector using one-hot encoding, resulting in $5,382$ features in total. A target value of 1 means the user has used the app under the context.  

\textbf{2. MovieLens}. This is the $Full$ version of the latest MovieLens data published by GroupLens~\cite{MovielensData}. As this work concerns higher-order interactions between features, we study the task of personalized tag recommendation rather than collaborative filtering~\cite{He:WWW2017} that considers the second-order interactions only. The tagging part of the data includes $668,953$ tag applications of $17,045$ users on $23,743$ items with $49,657$ distinct tags. We converted each tag application (\ie user ID, movie ID and tag) to a feature vector, resulting in $90,445$ features in total. A target value of 1 means the user has assigned the tag to the movie. \\\vspace{-5pt}

As both original datasets contain positive instances only (\ie all instances have target value 1), we sampled two negative instances to pair with one positive instance to ensure the generalization of the predictive model. 
For each log of Frappe, we randomly sampled two apps that the user has not used in the context; for each tag application of MovieLens, we randomly sampled two tags that the user has not assigned to the movie. 
Each negative instance is assigned to a target value of $-1$. Table~\ref{tab:dataset} summarizes the statistics of the final evaluation datasets. 

\begin{table}[t]
	\begin{center}
		\caption{\textbf{Statistics of the evaluation datasets.}}
		\vspace{-10pt}
		\label{tab:dataset}
		\begin{tabular}{ | l | c | c | c | c | }
			\hline
			\textbf{Dataset} & \textbf{Instance\#} & \textbf{Feature\#} & \textbf{User\#} & \textbf{Item\#} \\ \hline
			Frappe	& 288,609 & $5,382$ & 957 & 4,082\\ \hline
			MovieLens	& 2,006,859	& $90,445$ & 17,045 & 23,743 \\ \hline
		\end{tabular}
		\vspace{-15pt}
	\end{center}
\end{table}

\subsubsection{\textbf{Evaluation Protocols}} We randomly split the dataset into training (70\%), validation (20\%), and test (10\%) sets. The validation set was used for tuning hyper-parameters and the final performance comparison was conducted on the test set.
The study of NFM properties (\ie the answering of RQ1 and RQ2) was performed on the validation set, which can also reflect how we choose the optimal hyper-parameters. 
To evaluate the performance, we adopted \textit{root mean square error} (RMSE), where a lower RMSE score indicates a better performance. 
Note that RMSE has been widely used for evaluating regression tasks such as recommendation with explicit ratings~\cite{cao2017tois,fastFM} and click-through rate prediction~\cite{ImportanceFM}. 
We rounded up the prediction of each model to $1$ or $-1$ if it was out of the range. The one-sample paired t-test was performed to judge the statistical significance where necessary. 

\subsubsection{\textbf{Baselines}} 
We implemented NFM using TensorFlow\footnote{Codes are available at  \url{https://github.com/hexiangnan/neural_factorization_machine}}. We compared with the following competitive embedding-based models that are specifically designed for sparse data prediction:

- \textbf{LibFM}~\cite{libFM}. This is the official implementation\footnote{\url{http://www.libfm.org/}} of FM released by Rendle. It has shown strong performance for personalized tag recommendation and context-aware prediction~\cite{fastFM}. We used the SGD learner for a fair comparison with other methods which were all optimized with SGD (or its variants). 

- \textbf{HOFM}. This is the TensorFlow implementation\footnote{\url{https://github.com/geffy/tffm}} of higher-order FM, as described in \cite{FM}. We experimented with order size 3, since the MovieLens data concerns the ternary relationship between users, movies and tags.

- \textbf{Wide\&Deep}~\cite{WideDeep}. As introduced in Section~\ref{sec:dnn}, the deep part first concatenates feature embeddings, followed by a MLP to model feature interactions. As the structure of a DNN is difficult to be fully tuned, we used the same structure as reported in their paper, which has three layers with size $1024, 512$ and $256$, respectively. While the wide part (which is a linear regression model) is subjected to design to incorporate cross features, 
we used the raw features only for a fair comparison with FM and NFM. 

- \textbf{DeepCross}~\cite{DeepCross}. It applies a multi-layered residual network on the concatenation of feature embeddings for learning feature interactions. We used the same structure as reported in their paper, which stacks 5 residual units (each unit has two layers) with the hidden dimension $512, 512, 256, 128$ and $64$, respectively. 

\subsubsection{\textbf{Parameter Settings.}}
To fairly compare models' capability, we learned all models by optimizing the square loss (Equation (\ref{eq:squared_loss})). The learning rate was searched in $[0.005, 0.01, 0.02, 0.05]$ for all methods. 
To prevent overfitting, we tuned the $L_2$ regularization for linear models LibFM and HOFM in $[1e^{-6},5e^{-6},1e^{-5},...,1e^{-1}]$, and the dropout ratio for neural network models Wide\&Deep, DeepCross and NFM in $[0,0.1,0.2,...,0.9]$. 
Note that we found that dropout can well regularize the hidden layers of Wide\&Deep and NFM; however it did not work well for the residual units of DeepCross. 
Besides LibFM that optimized FM with the vanilla SGD, all other methods were optimized with mini-batch Adagrad~\cite{Adagrad}, where the batch size was set to 128 for Frappe and 4096 for MovieLens. Note that the batch size was selected by considering both training time and convergence rate, as a larger batch size usually led to faster training per epoch but slower convergence. 
For all methods, the early stopping strategy was performed, where we stopped training if the RMSE on validation set increased for 4 successive epochs. 
Without special mention, we show the results of embedding size 64, and more results of larger embedding sizes are shown in Section~\ref{ss:performance_comparison}.

\subsection{Study of Bi-Interaction Pooling (RQ1)}
We empirically study the Bi-Interaction pooling operation. 
To avoid other components (\eg hidden layers) affecting the analysis, we study the NFM-0 model that directly projects the output of Bi-Interaction pooling to prediction score with no hidden layer. As discussed in Section~\ref{ss:nfm_generalizes_fm}, NFM-0 is identical to FM as the trainable $\textbf{h}$ does not impact model's expressiveness. We first compare dropout with traditional $L_2$ regularization for preventing model overfitting, and then explore the impact of batch normalization. 

\subsubsection{\textbf{Dropout Improves Generalization}}
\label{ss:exper_dropout}
\begin{figure}[t]
	\centering
	\begin{subfigure}[b]{0.24\textwidth}
		\centering
		\includegraphics[width=\textwidth]{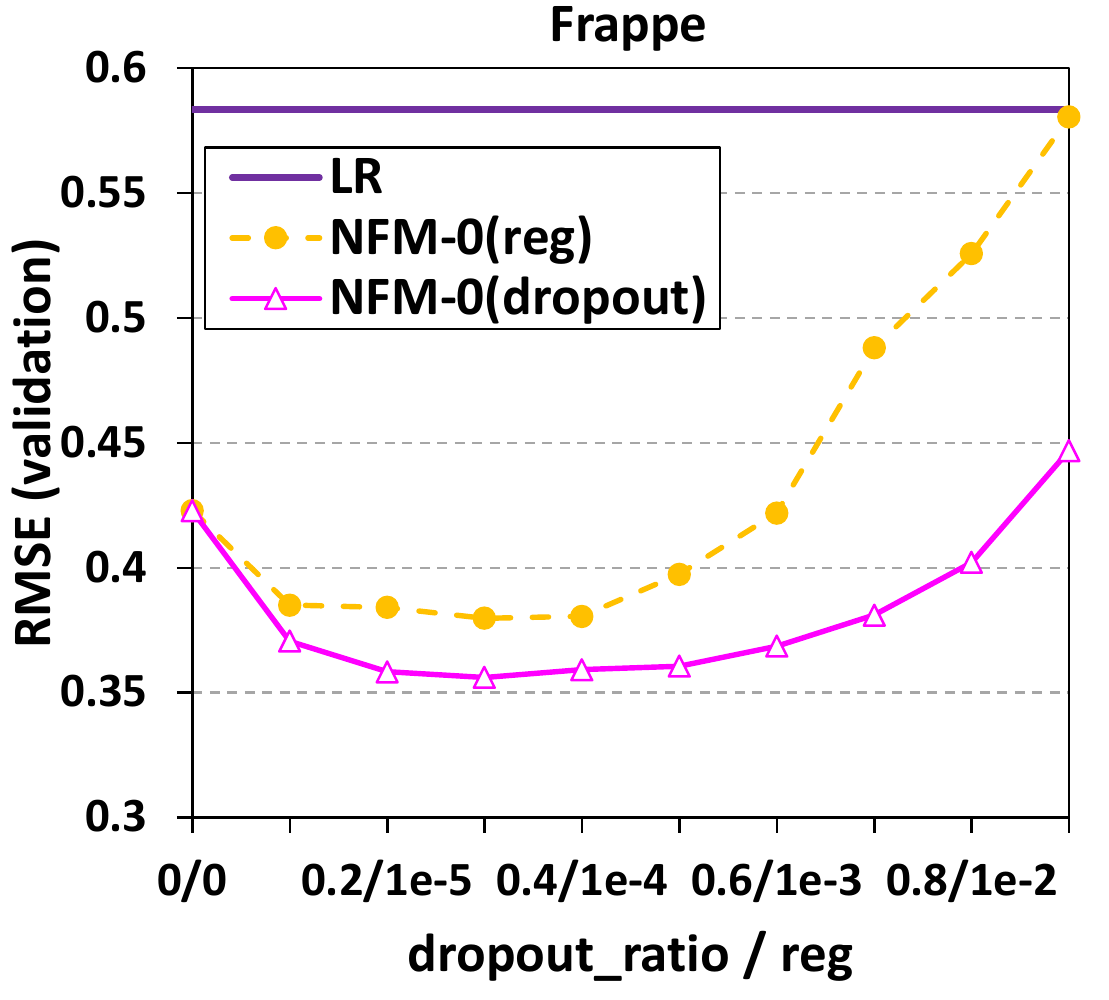}
		\vspace{-18pt}
		\label{fig:dropout_reg_frappe}
	\end{subfigure} \hspace{-7pt}
	\begin{subfigure}[b]{0.24\textwidth}
		\centering
		\includegraphics[width=\textwidth]{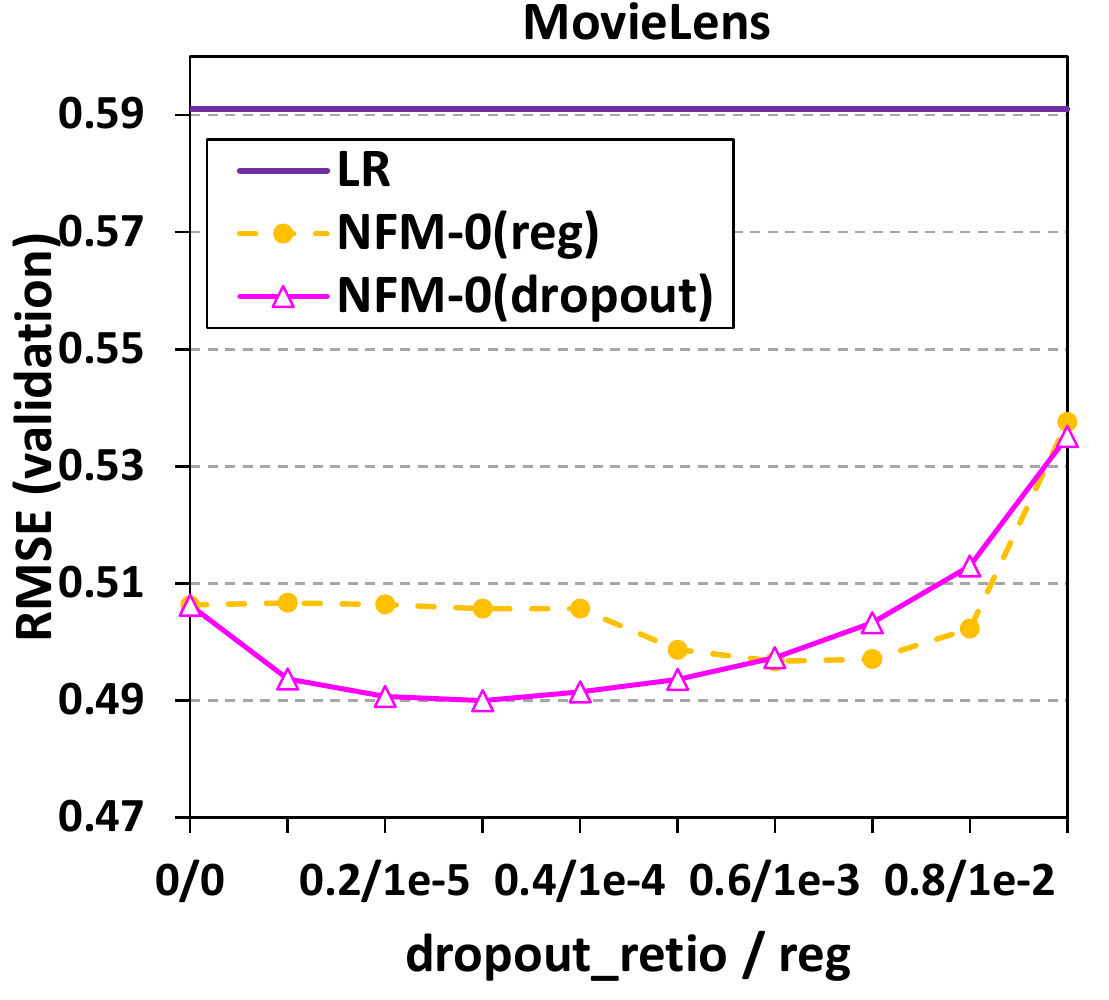}
		\vspace{-18pt}
		\label{fig:dropout_reg_ml}
	\end{subfigure} \hspace{-7pt}
	\caption{Validation error of NFM-0 \wrt dropout on the Bi-Interaction layer and $L_2$ regularization on embeddings.}
	\vspace{-15pt}
	\label{fig:dropout_reg}
\end{figure}

Figure~\ref{fig:dropout_reg} shows the validation error of NFM-0 \wrt dropout ratio on the Bi-Interaction layer and $L_2$ regularization on feature embeddings. The performance of linear regression (LR) is also shown for benchmarking the performance of prediction that does not consider feature interactions. 
First, LR leads to very poor performance, highlighting the importance of modelling interactions between sparse features for prediction. 
Second, we see that both $L_2$ regularization and dropout can well prevent overfitting and improve NFM-0's generalization to unseen data. 
Between the two strategies, dropout offers better performance. Specifically, on Frappe, using a dropout ratio of 0.3 leads to a lowest validation error of $0.3562$, which is significantly better than that of $L_2$ regularization $0.3799$.
One reason might be that enforcing $L_2$ regularization only suppresses the values of parameters in each update numerically, while using dropout can be seen as ensembling multiple sub-models~\cite{srivastava2014dropout}, which can be more effective. 
Considering the genericity of FM that subsumes many factorization models, we believe this is a new interesting finding, meaning that dropout can also be an effective strategy to address overfitting of linear latent-factor models. 

\begin{figure}[t]
	\centering
	\begin{subfigure}[b]{0.22\textwidth}
		\centering
		\includegraphics[width=\textwidth]{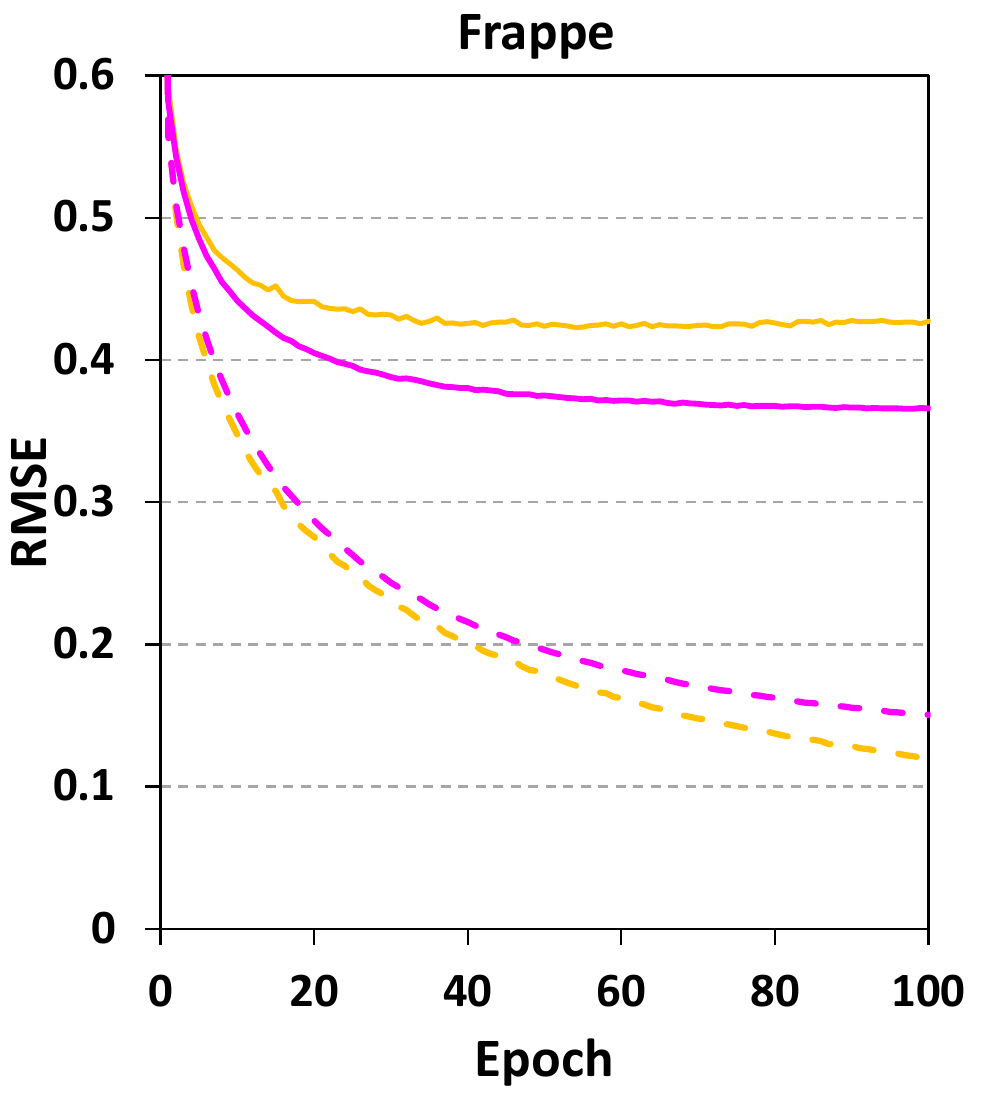}
		\vspace{-20pt}
		\label{fig:dropout_converge_frappe}
	\end{subfigure} 
	\begin{subfigure}[b]{0.22\textwidth}
		\centering
		\includegraphics[width=\textwidth]{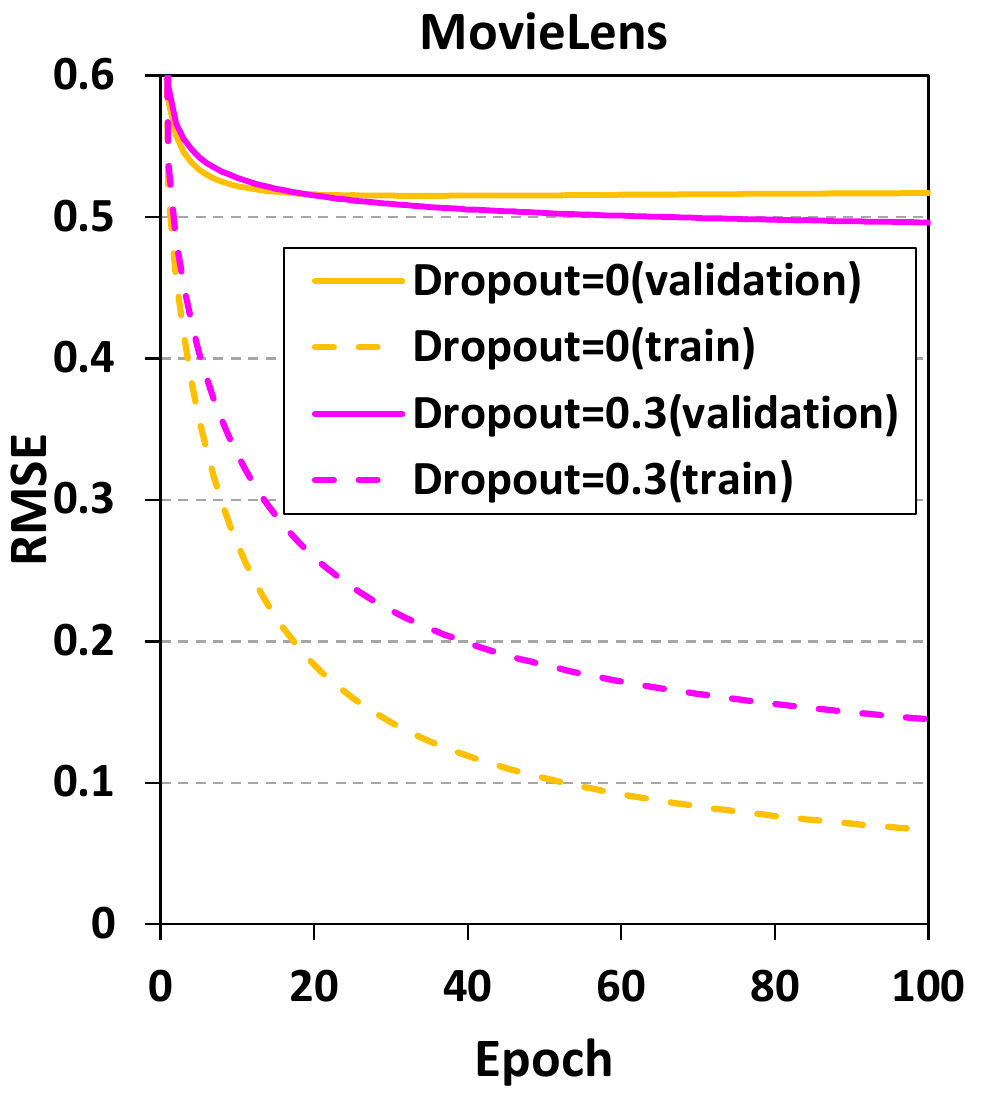}
		\vspace{-20pt}
		\label{fig:dropout_converge_ml}
	\end{subfigure} 
	\caption{Training and validation error of each epoch of NFM-0 with and without dropout on Bi-Interaction layer.}
	\vspace{-10pt}
	\label{fig:dropout_converge}
\end{figure}

To be more clear about the effect of dropout, we show the training and validation error of each epoch of NFM-0 with and without dropout in Figure \ref{fig:dropout_converge}. Both datasets show that with a dropout ratio of 0.3, although the training error is higher, the validation error becomes lower. This demonstrates the ability of dropout in preventing overfitting and as such, better generalization can be achieved. 

\subsubsection{\textbf{Batch Normalization Speeds up Training}} 
\begin{figure}[t]
	\centering
	\begin{subfigure}[b]{0.22\textwidth}
		\centering
		\includegraphics[width=\textwidth]{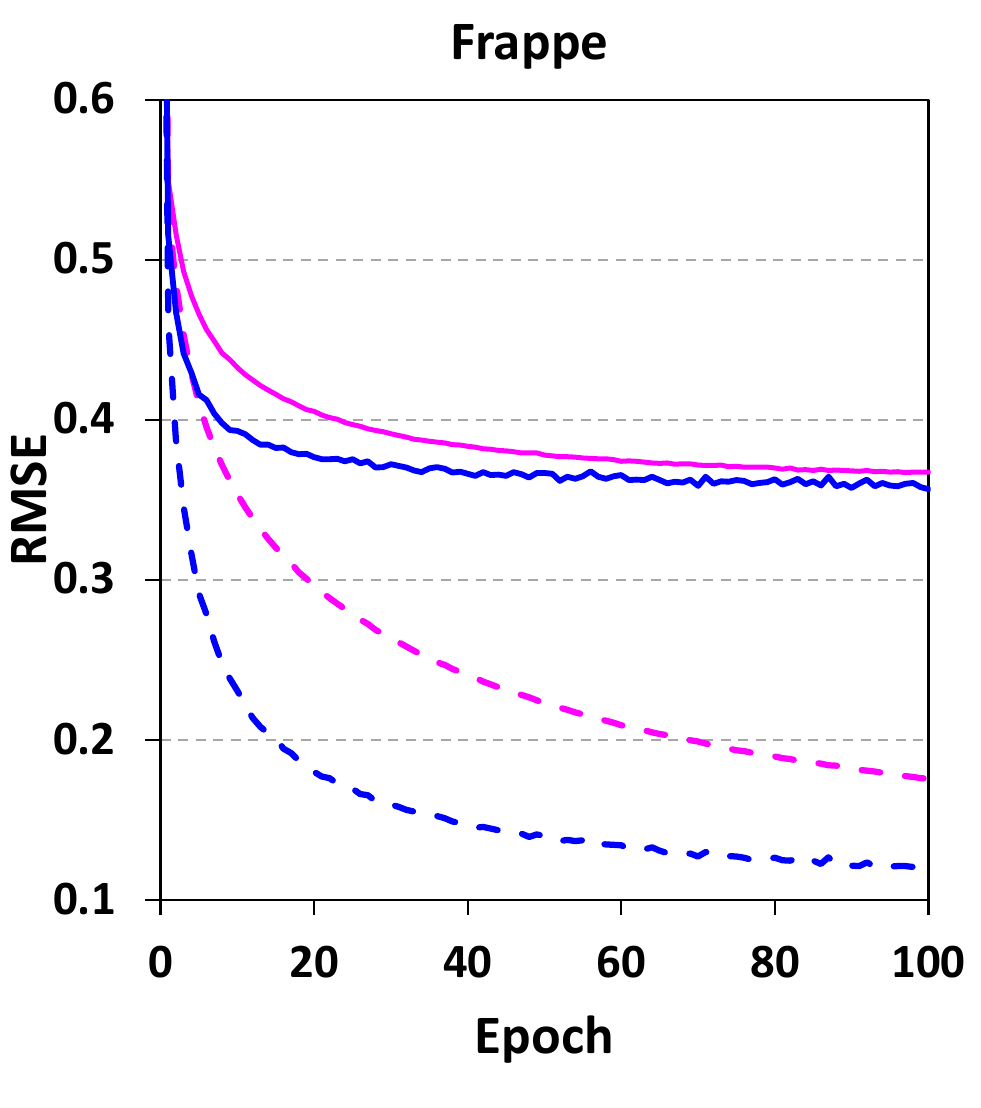}
		\vspace{-24pt}
		\label{fig:batchnorm_frappe}
	\end{subfigure} 
	\begin{subfigure}[b]{0.22\textwidth}
		\centering
		\includegraphics[width=\textwidth]{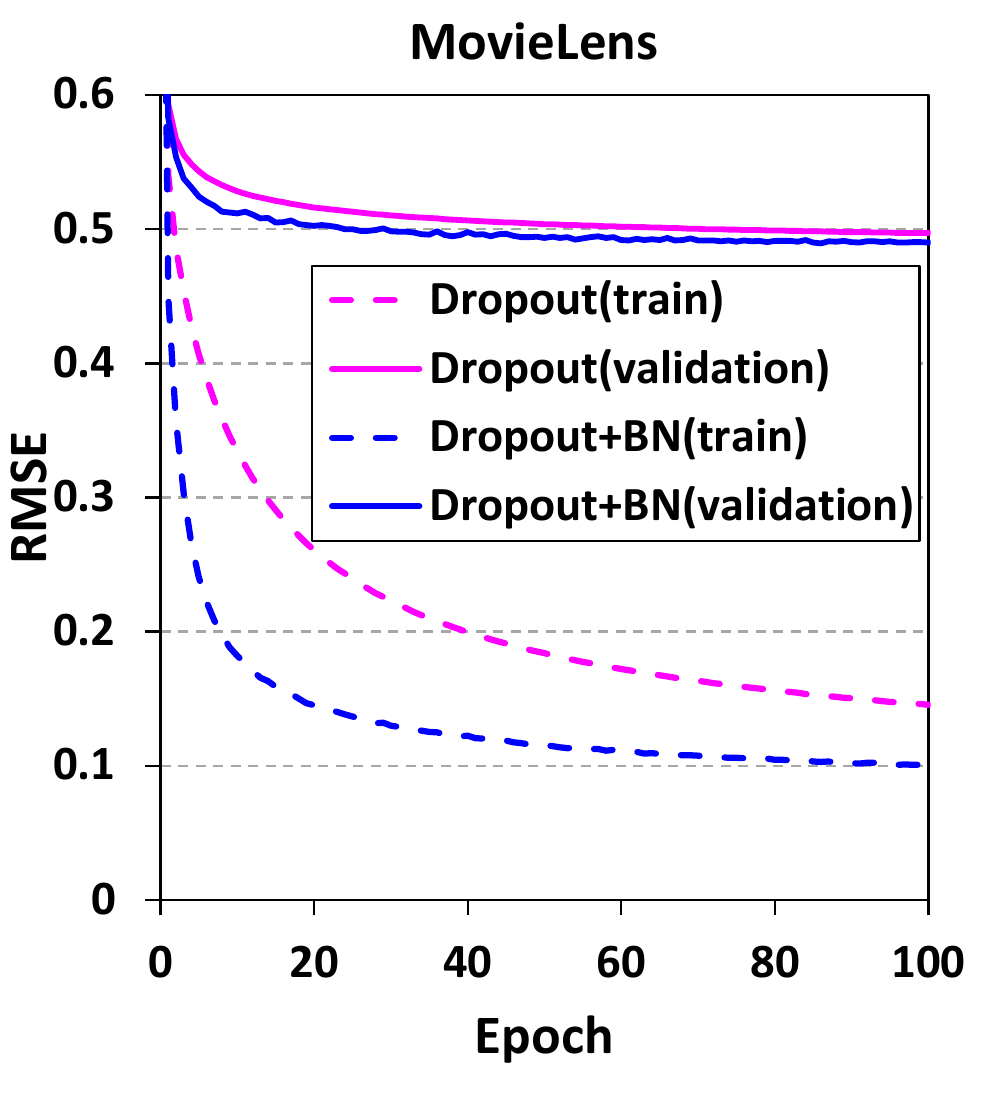}
		\vspace{-24pt}
		\label{fig:batchnorm_ml}
	\end{subfigure}
	\caption{Training and validation error of each epoch of NFM-0 with and without BN on the Bi-Interaction layer.}
	\vspace{-10pt}
	\label{fig:batchnorm}
\end{figure}

Figure~\ref{fig:batchnorm} shows the training and validation error of each epoch of NFM-0 with and without BN on the Bi-Interaction layer. The dropout is enabled with a ratio of 0.3, and the learning rate is set to 0.02. Focusing on the training error, we can see that BN leads to a faster convergence; on Frappe, when BN is applied, the training error of epoch 20 is even lower than that of epoch 60 without BN; and the validation error indicates that the lower training error is not overfitting --- in fact, \cite{cvpr16best,ioffe2015batch} showed that by addressing the internal covariate shift with BN, the model's generalization ability can be improved. Our result also verifies this point, where using BN leads to slight improvement (although the improvement is not statistically significant). 
Furthermore, we notice that BN makes the learning less stable, as evidenced by the larger performance fluctuation of blue lines. 
This is caused by our use of dropout and BN together, as randomly dropping neurons can change the input distribution normalized by BN. It is an interesting direction to effectively combine BN and dropout. 

\subsection{Impact of Hidden Layers (RQ2)}

\begin{figure}[t]
	\centering
	\begin{subfigure}[b]{0.24\textwidth}
		\centering
		\includegraphics[width=\textwidth]{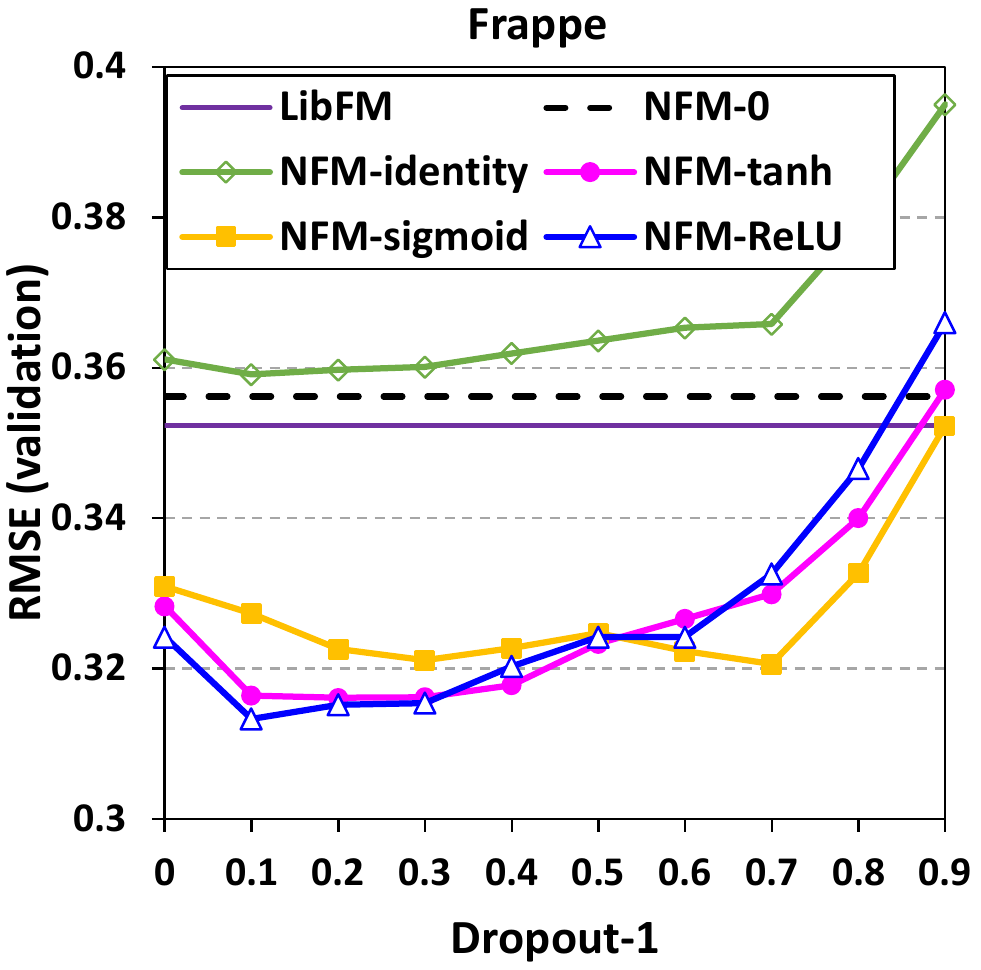}
		\vspace{-20pt}
		\label{fig:layers_frappe}
	\end{subfigure} \hspace{-7pt}
	\begin{subfigure}[b]{0.24\textwidth}
		\centering
		\includegraphics[width=\textwidth]{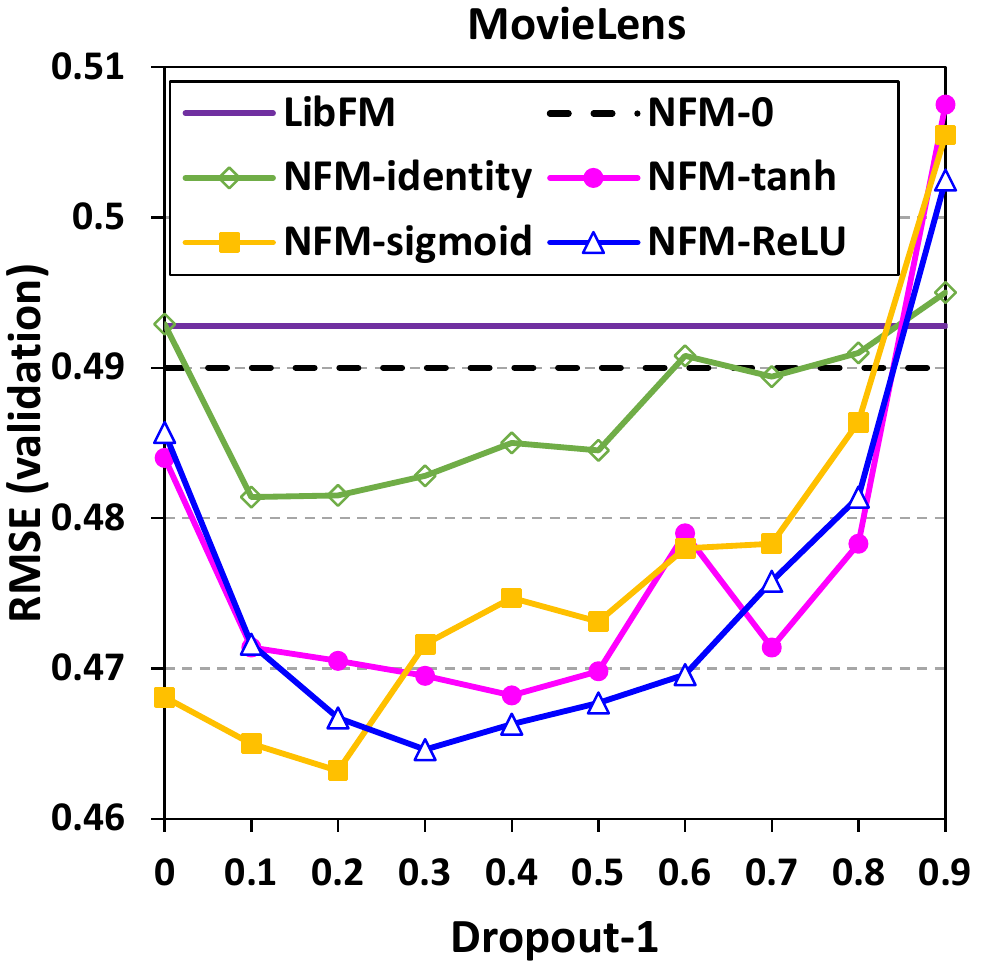}
		\vspace{-20pt}
		\label{fig:layers_ml}
	\end{subfigure} \hspace{-7pt}
	\caption{Validation error of LibFM, NFM-0 and NFM with different activation functions on the first hidden layer.}
	\vspace{-10pt}
	\label{fig:layers}
\end{figure}

The hidden layers of NFM play a pivotal role in capturing higher-order interactions between features. 
To explore the impact, we first add one hidden layer above the Bi-Interaction layer and slightly overuse the name NFM to indicate this specific model. 
To ensure the same model capability with NFM-0, we set the size of hidden layer the same as the embedding size.

Figure~\ref{fig:layers} shows the validation error of NFM \wrt different activation functions and dropout ratios for the hidden layer. The performance of LibFM and NFM-0 are also shown for benchmarking purposes. 
First and foremost, we observe that by using non-linear activations, NFM's performance is improved with a large margin --- compared to NFM-0 which has a similar performance with LibFM, the relative improvement is $11.3\%$ and $5.2\%$ for Frappe and MovieLens, respectively. 
This highlights the importance of modelling higher-order feature interactions for quality prediction. 
Among the different non-linear activation functions, there is no obvious winner. 
Second, when we use the identity function as the activation function, \ie the hidden layer performs a linear transformation, NFM does not perform that well. 
This provides evidence to the necessity of learning higher-order feature interactions with non-linear functions. 

\begin{figure}[t]
	\centering
	\begin{subfigure}[b]{0.22\textwidth}
		\centering
		\includegraphics[width=\textwidth]{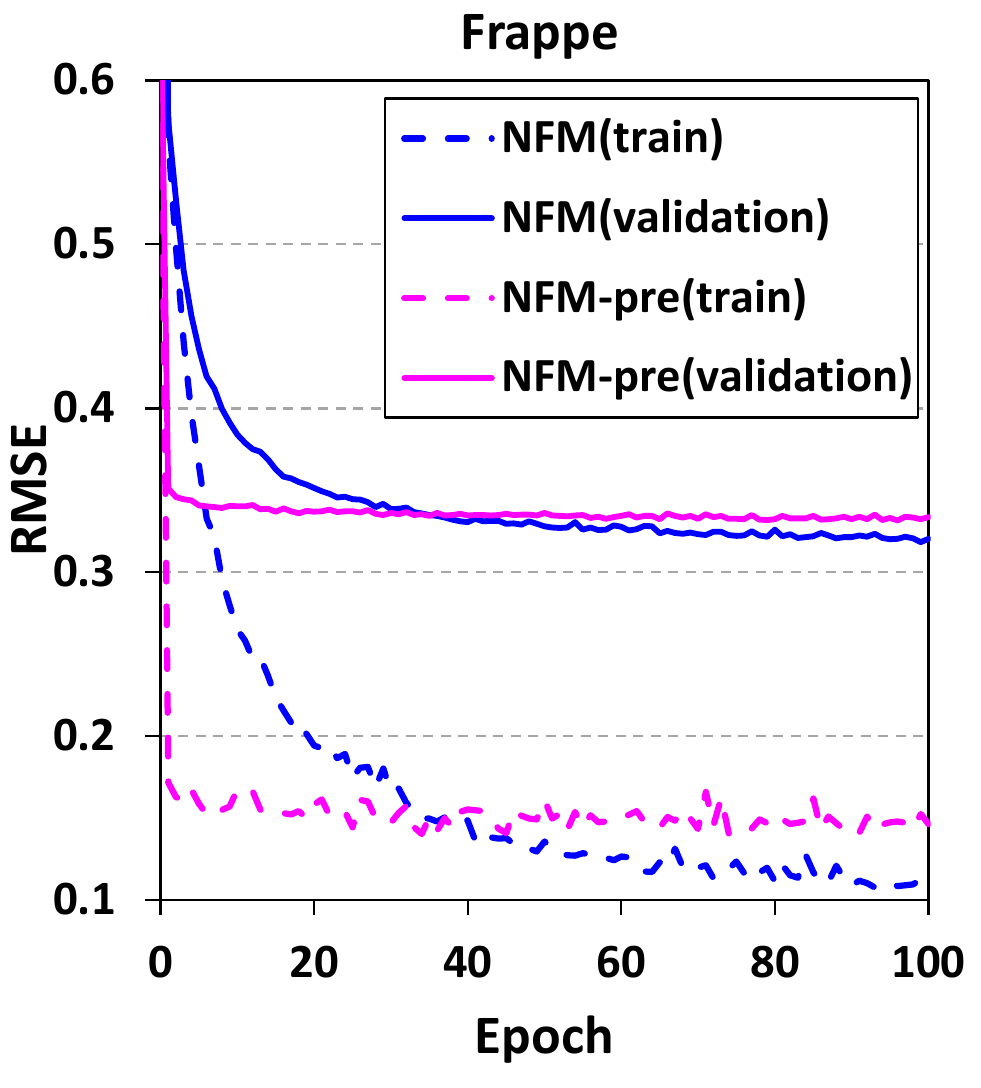}
		\vspace{-22pt}
		\label{fig:pretrain_frappe}
	\end{subfigure} 
	\begin{subfigure}[b]{0.22\textwidth}
		\centering
		\includegraphics[width=\textwidth]{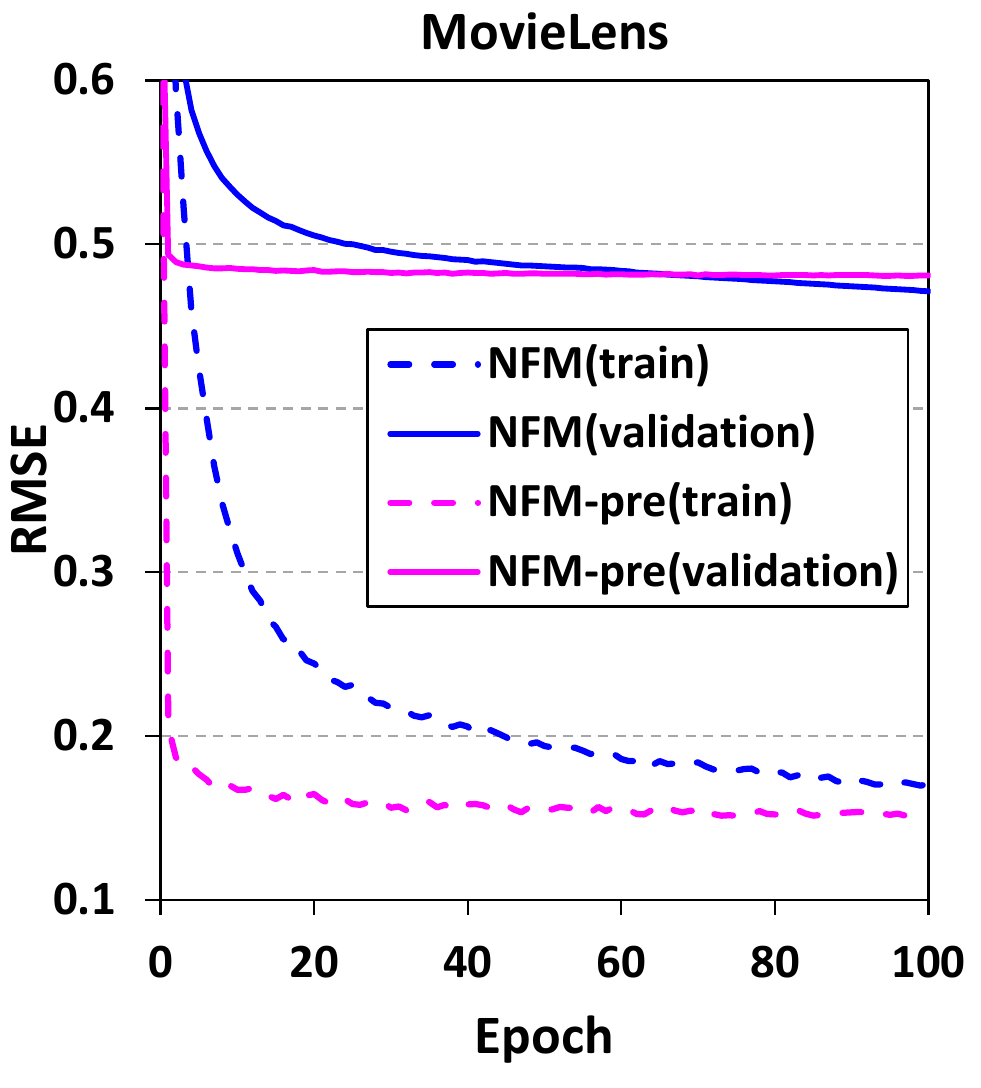}
		\vspace{-22pt}
		\label{fig:pretrain_ml}
	\end{subfigure} 
	\caption{Training and validation error of each epoch of NFM-1 with and without pre-training.}
	\vspace{-12pt}
	\label{fig:pretrain}
\end{figure}

To see whether a deeper NFM can further improve the performance, we stack more ReLU layers above the Bi-Interaction layer. 
As it is computationally expensive to tune the size and dropout ratio for each hidden layer separately, we use the same setting for all layers and tune them the same way as NFM-1. 
As can be seen from Table~\ref{tab:layers}, when we stack more layers, the performance is not further improved, and best performance is when we use one hidden layer only. 
We have also explored other designs for hidden layers,
such as the tower structure and residual units, however, the performance is still not improved. 
We think the reason is because the Bi-Interaction layer has encoded informative second-order feature interactions, and based on which, a simple non-linear function is sufficient to capture higher-order interactions. 
To verify this, we replaced the Bi-Interaction layer with concatenation (which leads to the same architecture as Wide\&Deep), and found that the performance can be gradually improved with more hidden layers (up to three); however, the best performance achievable is still inferior to that of NFM-1.
This demonstrates the value of using a more informative operation for low-level layers, which can ease the burden of higher-level layers for learning meaningful information. As a result, a deep structure becomes not necessarily required.

\begin{table}[h]
	\vspace{-10pt}
	\begin{center}
		\caption{NFM \wrt different number of hidden layers.}
		\vspace{-10pt}
		\label{tab:layers}
		\begin{tabular}{| l | c | c |} \hline
			\textbf{Methods} & \textbf{Frappe} & \textbf{MovieLens} \\ \hline\hline
			NFM-0 	& 0.3562  & 0.4901 \\ \hline
			NFM-1 	& \textbf{0.3133}  & \textbf{0.4646} \\ \hline
			NFM-2 	& 0.3193  & 0.4681 \\ \hline
			NFM-3 	& 0.3219 & 0.4752 \\ \hline
			NFM-4 	& 0.3202  & 0.4703 \\ \hline
		\end{tabular}
	\end{center}
	\vspace{-10pt}
\end{table}


\begin{table*}[t]
	\small
	\begin{center}
		\caption{Test error and number of trainable parameters for different methods on latent factors 128 and 256. M denotes ``million''; $*$ and $**$ denote the statistical significance for $p<0.05$ and $p<0.01$, respectively, compared to the best baseline. }
		\vspace{-10pt}
		\label{tab:performance}
		\begin{tabular}{| l | c | c | c | c | c | c | c | c |} \hline
			& \multicolumn{4}{c|}{\textbf{Frappe}} & \multicolumn{4}{c|}{\textbf{MovieLens}} \\ \hline 
			& \multicolumn{2}{c|}{\textbf{Factors=128}} & \multicolumn{2}{c|}{\textbf{Factors=256}} & \multicolumn{2}{c|}{\textbf{Factors=128}} & \multicolumn{2}{c|}{\textbf{Factors=256}}\\ \hline 
			\textbf{Method} & \textbf{Param\#} & \textbf{RMSE} & \textbf{Param\#} & \textbf{RMSE} & \textbf{Param\#} & \textbf{RMSE} & \textbf{Param\#} & \textbf{RMSE} \\ \hline\hline
			LibFM~\cite{libFM} 	& 0.69M  & 0.3437 & 1.38M  & 0.3385 & 11.67M  & 0.4793 & 23.24M  & 0.4735 \\ \hline
			HOFM & 1.38M  & 0.3405 & 2.76M & 0.3331 & 23.24M & 0.4752
			& 46.40M & 0.4636 \\ \hline\hline
			Wide\&Deep~\cite{WideDeep} &  2.66M  & 0.3621 & 4.66M & 0.3661 & 12.72M & 0.5323 & 24.69M  & 0.5313 \\ \hline
			Wide\&Deep~(pre-train) &  2.66M  & 0.3311 & 4.66M & 0.3246  & 12.72M & 0.4595 & 24.69M  & 0.4512 \\ \hline \hline
			DeepCross~\cite{DeepCross}	& 4.47M  & 0.4025 & 8.93M & 0.4071  & 12.71M & 0.5885 & 25.42M  & 0.5907 \\ \hline
			DeepCross~(pre-train)	& 4.47M  & 0.3388 & 8.93M & 0.3548  & 12.71M & 0.5084 & 25.42M  & 0.5130  \\ \hline \hline
			\textbf{NFM} & \textbf{0.71M} & $\textbf{0.3127}^{**}$ & \textbf{1.45M}  & $\textbf{0.3095}^{**}$ & \textbf{11.68M} & $\textbf{0.4557}^*$ & \textbf{23.31M}  & $\textbf{0.4443}^*$\\ \hline
		\end{tabular}
	\end{center}
	\vspace{-8pt}
\end{table*}

\begin{figure*}[t]
	\centering
	\begin{subfigure}[b]{0.36\textwidth}
		\centering
		\includegraphics[width=\textwidth]{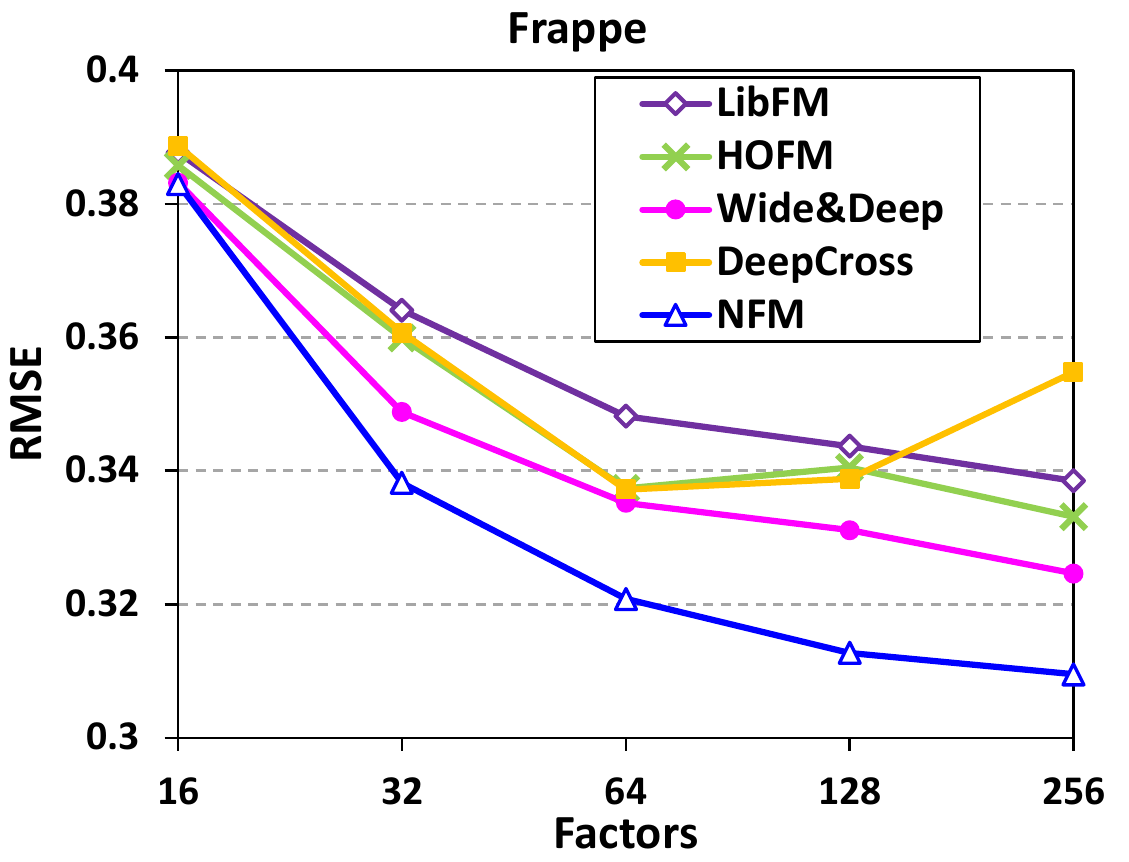}
		\vspace{-15pt}
		\label{fig:performance_frappe}
	\end{subfigure} \hspace{+55pt}
	\begin{subfigure}[b]{0.36\textwidth}
		\centering
		\includegraphics[width=\textwidth]{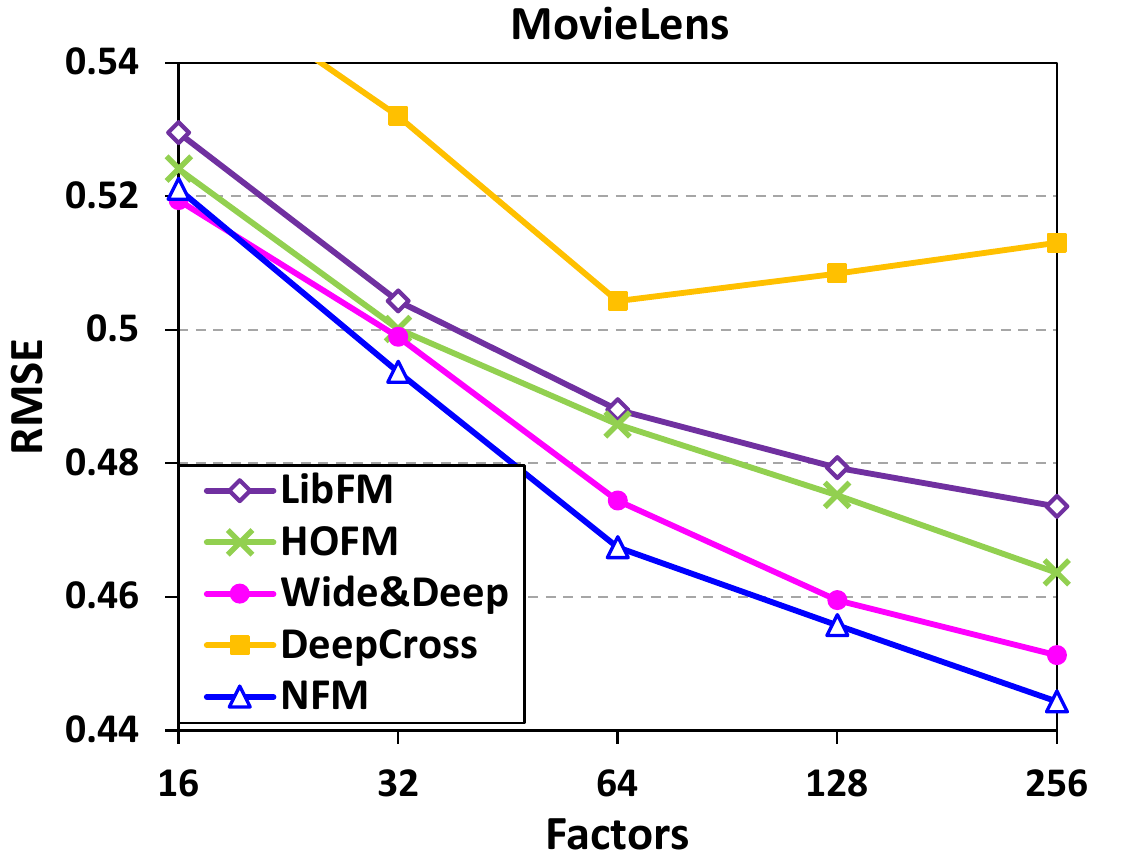}
		\vspace{-15pt}
		\label{fig:performance_ml}
	\end{subfigure} \hspace{+7pt}
	\caption{Performance comparison on the test set \wrt different embedding sizes. LibFM, HOFM and HOFM are trained from random initialization; Wide\&Deep and DeepCross are pre-trained with FM feature embeddings.}
	\vspace{-10pt}
	\label{fig:performance}
\end{figure*}

\subsubsection{\textbf{Pre-training Speeds up Training}}
It is known that parameter initialization can greatly affect the convergence and performance of DNNs~\cite{erhan2010pretrain,He:WWW2017}, since gradient-based methods can only find local optima for DNNs. 
As have been shown in Section~\ref{ss:difficulty_dnn}, initializing with feature embeddings learned by FM can significantly enhance Wide\&Deep and DeepCross. Now the question arises, how does pre-training impact NFM? 

Figure~\ref{fig:pretrain} shows the state of each epoch of NFM-1 with and without pre-training.
First, we can see that by using FM embeddings as pre-training, NFM exhibits extremely fast convergence --- on both datasets, with 5 epochs only, the performance is on par with 40 epochs of NFM that is trained from scratch (with BN enabled).
Second, we find that pre-training does not improve NFM's final performance, and a random initialization can achieve a result that is slightly better than that with pre-training. 
This demonstrates the robustness of NFM, which is relatively insensitive to parameter initialization. 
In contrast to the huge impact of pre-training on Wide\&Deep and DeepCross (\cf Figure~\ref{fig:difficulty}) that improves both their convergence and final performance, we draw the conclusion that NFM is much easier to train and optimize, which is due largely to the informative and effective Bi-Interaction pooling operation. \vspace{-5pt}

\subsection{Performance Comparison (RQ3)}
\label{ss:performance_comparison}
We now compare with state-of-the-art methods. 
For NFM, we use one hidden layer with ReLU as the activation function, since the baselines DeepCross and Wide\&Deep also choose ReLU in their original papers. Note that the most important hyper-parameter for NFM is the dropout ratio, which we use $0.5$ for the Bi-Interaction layer and tune the value for the hidden layer. 

Figure~\ref{fig:performance} plots the test RMSE \wrt different number of latent factors (\ie embedding sizes), where Wide\&Deep and DeepCross are pre-trained with FM to better explore the two methods. 
Table~\ref{tab:performance} shows the concrete scores obtained on factors 128 and 256, and the number of model parameters of each method. The scores of Wide\&Deep and DeepCross without pre-training are also shown. We have the following three key observations. 

First and foremost, NFM consistently achieves the best performance on both datasets with the fewest model parameters besides FM. This demonstrates the effectiveness and rationality of NFM in modelling higher-order and non-linear feature interactions for prediction with sparse data. 
The performance is followed by Wide\&Deep, which uses a 3-layer MLP to learn feature interactions. We have also tried deeper layers for Wide\&Deep, however the performance has not been improved. This further verifies the utility of using the informative Bi-Interaction pooling in the low level. 

Second, we observe that HOFM shows slight improvement over FM with $1.45\%$ and $1.04\%$ average improvement on Frappe and MovieLens, respectively. This sheds light on the limitation of FM that models only the second-order feature interactions, and thus the usefulness of modelling higher-order interactions. Meanwhile, the large performance gap between HOFM and NFM reflects the value of modelling higher-order interactions in a non-linear way, since HOFM models higher-order interactions linearly and uses much more parameters than NFM. 

Lastly, the relatively weak performance of DeepCross reveals that deeper learnings are not always better, as DeepCross is the deepest method among all baselines that utilizes a 10-layer network. 
On Frappe, DeepCross only achieves a comparable performance with the shallow FM model, while it underperforms FM significantly on MovieLens. 
We believe that the reasons are due to optimization difficulties and overfitting (as evidenced by the worse performance on factors 128 and 256). \\\vspace{-8pt}

%
%
%

\section{Conclusion and Future Work}
\label{sec:conclusion}

In this work, we proposed a novel neural network model NFM, which brings together the effectiveness of linear factorization machines with the strong representation ability of non-linear neural networks for sparse predictive analytics.
The key of NFM's architecture is the newly proposed Bi-Interaction  operation, based on which we allow a neural network model to learn more informative feature interactions at the lower level. 
Extensive experiments on two real-world datasets show that with one hidden layer only, NFM significantly outperforms FM, higher-order FM, and state-of-the-art deep learning approaches Wide\&Deep and DeepCross. 

The work represents the first step towards bridging the gap between linear models and deep learning. Linear models, such as various factorization methods, have shown to be effective for many IR and DM tasks and are easy to interpret.
However, their limited expressiveness may hinder the performance when modelling real-world data with complex inherent patterns. 
While deep learning models have exhibited great expressive power and yielded immense success on speech processing and computer vision, their performance is still unsatisfactory for IR tasks, such as collaborative filtering~\cite{He:WWW2017}. In our view, one reason is that most data of IR and DM tasks are naturally sparse; and to date, there still lacks effective deep learning solutions for prediction with sparse data. 
By connecting neural networks with FM --- one of the most powerful linear models for supervised learning --- we are able to design a simple yet effective deep learning solution for sparse data prediction. 

With recent developments on GPU platforms, it is not technically difficult to build very deep models with hundreds or even thousands of layers~\cite{cvpr16best}. However, deeper models do not necessarily lead to better results, since deeper models are less transparent and more difficult to optimize and tune. 
As such, we expect future research on deep learning for IR should focus more on designing better neural components or architectures for specific tasks, rather than relying on deeper models for minor improvements. 
Our proposed Bi-Interaction pooling is an effective neural component for sparse feature modelling, reducing the demand for deeper structure for quality prediction. 
In future, we will improve the efficiency of NFM by resorting to hashing techniques~\cite{DCF:2016,shen2017binary} to make it more suitable for large-scale applications and study its performance for other IR tasks, such as search ranking and targeted advertising. While this work endows FM with non-linearities from predictive model perspective, another viable solution for incorporating non-linearities is to extend the objective function with regularizers like the graph Laplacian~\cite{he2014popularity,meng_semi_supervised}. Lastly, we are interested in exploring the Bi-Interaction pooling for recurrent neural networks (RNNs) for sequential data modelling. 

\bibliographystyle{abbrv}\balance
\bibliography{proc-nopage}\balance
\end{document}